\pdfoutput=1
\documentclass[fleqn,10pt]{wlpeerj}

\makeatletter
\g@addto@macro\normalsize{%
  \setlength\abovedisplayskip{4pt plus 4pt minus 3pt}%
  \setlength\belowdisplayskip{4pt plus 4pt minus 3pt}%
  \setlength\abovedisplayshortskip{4pt plus 3pt minus 2pt}%
  \setlength\belowdisplayshortskip{4pt plus 3pt minus 2pt}%
}
\makeatother

\usepackage{threeparttable}
\usepackage{array}
\usepackage{xcolor}
\usepackage{graphicx}
\usepackage{comment}
\usepackage{amsmath,amssymb,amsfonts}  
\usepackage{algorithm}
\usepackage{algpseudocode}
\usepackage{float}
\usepackage{booktabs}
\usepackage{multirow}
\usepackage[hidelinks]{hyperref}

\title{GenDiff: A dose and anatomy aware diffusion model with structural prior refinement for low-dose CT reconstruction and generalization}
\author[1]{Md Imam Ahasan}
\author[2]{Guangchao Yang}
\author[3]{A. F. M. Abdun Noor}
\author[4]{Kah Ong Michael Goh}
\author[5]{S. M. Hasan Mahmud}
\author[6]{Md Mahfuzur Rahman}

\affil[1, 2, 6]{College of Computer Science, Chongqing University, Chongqing, China}
\affil[3, 5]{Department of Software Engineering, Daffodil International University, Bangladesh}
\affil[4]{Faculty of Information Science \& Technology, Multimedia University, Malaysia}

\corrauthor{Kah Ong Michael Goh, Guangchao Yang}{michael.goh@mmu.edu.my, gchao\_yang@cqu.edu.cn}

\begin{abstract}
Computed tomography (CT) is a critical imaging modality in clinical diagnosis but exposes patients to ionizing radiation. Reducing radiation dose is therefore essential; however, low-dose computed tomography (LDCT) introduces severe noise and structured artifacts that degrade diagnostic quality. Existing learning-based reconstruction methods are typically trained for fixed dose levels or specific anatomies, limiting their robustness under realistic clinical variability. We propose GenDiff, a generalizable diffusion-based framework for LDCT reconstruction. The proposed approach integrates a Dose-Anatomy Encoder to learn continuous acquisition-aware embeddings, a dose- and anatomy-conditioned cold diffusion backbone for iterative refinement, and a physics-consistency update mechanism to enforce adherence to the CT forward model. To further preserve anatomical fidelity under severe dose reduction, a Structural Prior Refinement Module (SPRM) is incorporated as a learned proximal operator, complemented by contextual error modulation for spatially adaptive correction. Experiments on multi-anatomy clinical datasets, including unseen ultra-low-dose conditions and out-of-distribution phantom and animal data, demonstrate that GenDiff achieves consistently strong performance. In particular, it improves peak signal-to-noise ratio (PSNR) by up to \(2.1\) dB and structural similarity index (SSIM) by up to \(0.035\) compared with state-of-the-art convolutional neural network (CNN) and diffusion-based methods.
\end{abstract}

\begin{document}

\flushbottom
\maketitle
\thispagestyle{empty}

\section*{Introduction}

Computed tomography (CT) is an indispensable imaging modality in modern clinical diagnosis that supports a wide range of applications from oncology and cardiovascular assessment to emergency medicine~\cite{camacho2025clinical}. Despite its diagnostic value, CT relies on ionizing X-ray radiation that is associated with non-negligible health risks including radiation-induced tissue damage and increased lifetime cancer risk~\cite{smith2009radiation, Sodickson2009RecurrentCC}. These concerns have driven sustained efforts to reduce radiation exposure in accordance with the as low as reasonably achievable (ALARA) principle~\cite{Shah2008ALARAIT}, while maintaining image quality sufficient for reliable clinical interpretation. In practice, low-dose CT (LDCT) acquisition is typically achieved through two complementary strategies by reducing the number of projection views (sparse-view CT) or lowering the X-ray tube current intensity (dose reduction)~\cite{Yan2011ACS}. Although both approaches effectively reduce patient dose, they inevitably amplify quantum noise and introduce structured artifacts into reconstructed images which substantially degrades diagnostic quality~\cite{Humphries2019ComparisonOD}. Fixing these issues with advanced reconstruction algorithms is still a major challenge in medical imaging. LDCT reconstruction is an ill-posed problem because the reduced and noisy measurements do not contain enough information to reliably recover high-quality images~\cite{Attivissimo2010ATT, Wang2020DeepLF}.
\begin{figure}[!ht]
    \centering
    \includegraphics[width=1\linewidth]{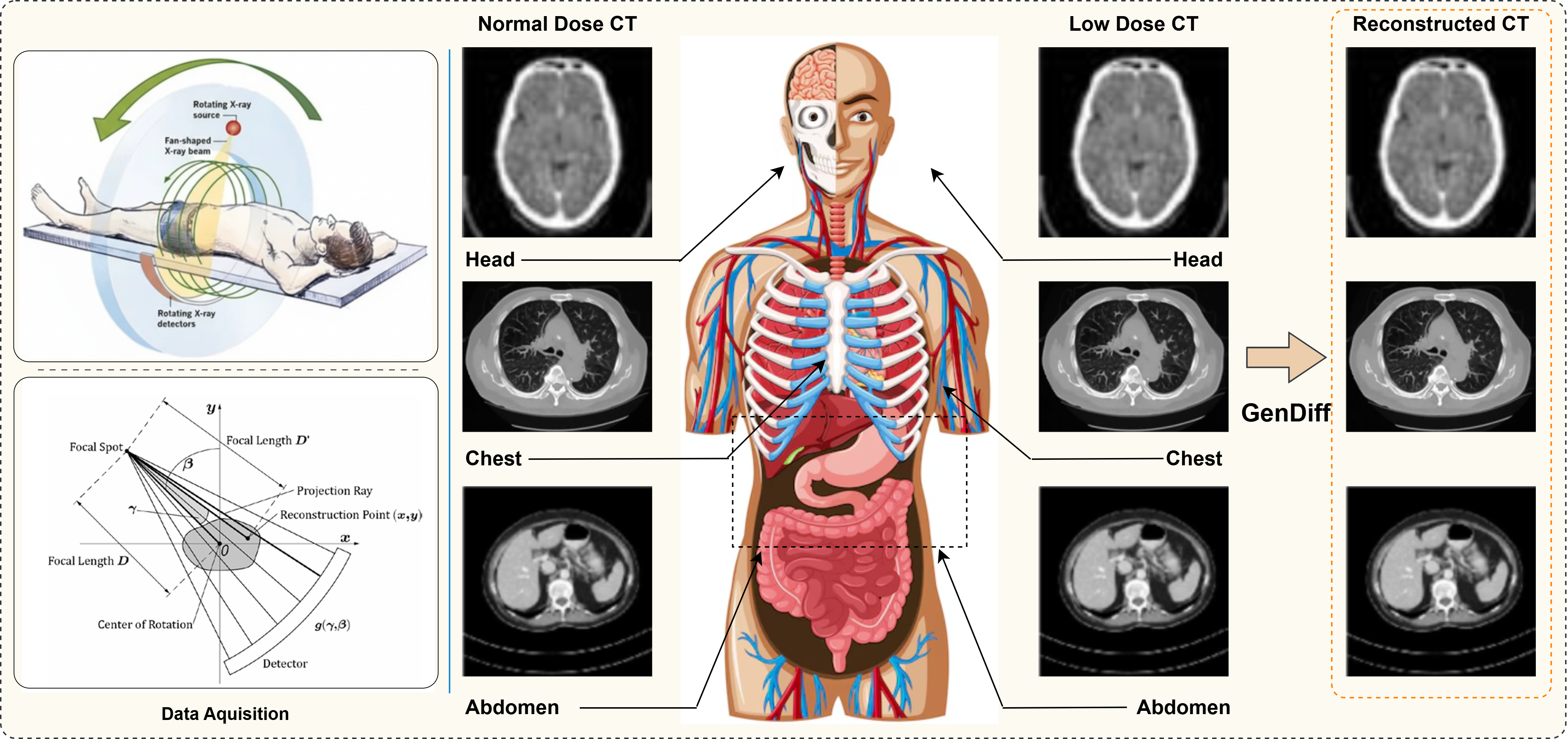}
    \caption{Overview of the proposed GenDiff framework for unified low-dose CT reconstruction. GenDiff integrates doseand anatomy-aware conditioning within a deterministic cold diffusion process to reconstruct high-quality CT images from low-dose inputs. The framework jointly models radiation dose and anatomical region, enforces physics consistency during iterative refinement, and produces robust reconstructions across multiple anatomies and unseen dose levels.}
    \label{fig:concept}
\end{figure}
In recent years, deep learning (DL) has  become as a powerful paradigm for LDCT image denoising and reconstruction and it delivers substantial improvements over traditional analytical methods~\cite{Shan2018CompetitivePO, Wang2020DeepLF}. Early methods mainly used supervised encoder-decoder networks. They were trained to reduce pixel-level differences between LDCT and normal-dose CT (NDCT) images. A well-known example is the residual encoder-decoder convolutional neural network (RED-CNN)~\cite{Chen2017LowDoseCW}. Subsequent extensions introduced parameter-dependent modeling to accommodate multiple geometries and dose levels~\cite{Xia2021CTRW}. Despite their effectiveness, these regression based methods often produce over smoothed reconstructions which leads to the loss of fine anatomical structures and clinically relevant texture details~\cite{Kim2019APC, Nagare2021ABL}. To mitigate this limitation, generative adversarial networks (GANs) have been explored to enhance perceptual realism by encouraging reconstructions that more closely resemble NDCT images~\cite{Kang2018CycleconsistentAD}. Representative methods include WGAN-VGG, which combines Wasserstein adversarial training with perceptual loss~\cite{Yang2017LowDoseCI}, and DU-GAN, which employs dual-domain discriminators to capture both global and local discrepancies~\cite{Huang2021DUGANGA}. However, GAN-based methods are notoriously difficult to train, often suffering from instability and sensitivity to architectural and optimization choices~\cite{Zhao2018BiasAG}. In parallel, classical iterative reconstruction techniques that integrate sinogram-domain measurements with image-domain priors, such as total variation minimization~\cite{Chambolle1997ImageRV}, dictionary learning~\cite{Xu2012LowDoseXC}, and low-rank regularization~\cite{Wu2018NonLocalLC}, are theoretically well-founded but remain computationally expensive and highly sensitive to hand-tuned parameters.

More recently, hybrid approaches have sought to combine the strengths of iterative reconstruction and deep learning. Unrolled optimization frameworks, including LEARN~\cite{Chen2017LEARNLE}, plug-and-play ADMM variants~\cite{He2019OptimizingAP}, and manifold-aware reconstruction networks~\cite{Xia2020MAGICMA}, explicitly embed physical measurement models within learnable architectures. While these methods improve reconstruction fidelity, their reliance on deep iterative pipelines substantially increases training complexity and inference time, limiting scalability and clinical practicality. In parallel, diffusion models have attracted growing attention as a class of generative models with strong theoretical foundations and impressive empirical performance~\cite{SohlDickstein2015DeepUL, Chung2022ImprovingDM}. By modeling data distributions through gradual perturbation and denoising, diffusion models combine the mode coverage of variational autoencoders with the high-fidelity synthesis of GANs~\cite{Song2020ScoreBasedGM, Dhariwal2021DiffusionMB, Yang2022DiffusionMA}. However, for CT reconstruction, generative fidelity alone is insufficient, as clinically reliable solutions must also satisfy strict consistency with the underlying X-ray measurement physics. Classical denoising diffusion probabilistic models (DDPMs) typically require hundreds to thousands of sampling steps, resulting in prohibitively high inference costs for time-sensitive medical imaging applications~\cite{Ho2020DenoisingDP, abdun2025geglunet, Gao2022CoCoDiffAC, Xia2022LowDoseCU}. Accelerated variants have been proposed to address this limitation, including improved DDPM formulations with fewer sampling steps~\cite{Nichol2021ImprovedDD} and fast solvers for LDCT denoising~\cite{Xia2022LowDoseCU}. Nevertheless, these approaches largely operate within Gaussian noise-based diffusion frameworks and primarily trade reconstruction quality for computational efficiency.

Cold diffusion has recently emerged as a generalized diffusion paradigm that replaces stochastic noise corruption with deterministic, task-specific degradation operators~\cite{Bansal2022ColdDI, Yen2022ColdDF}. By explicitly modeling realistic degradation processes, cold diffusion offers improved interpretability and flexibility. However, its performance critically depends on the accuracy of the learned restoration operator, and imperfect restoration can lead to error accumulation across diffusion steps, resulting in deviations from the ground-truth solution. Building on this paradigm, Gao \emph{et al.}\ proposed a contextual error-modulated diffusion model(CoreDiff) for LDCT denoising that leverages degradation-aware guidance~\cite{Gao2023CoreDiffCE}. In parallel, score-based diffusion methods have been extended to inverse problems by incorporating measurement consistency, manifold constraints, and prior information~\cite{Chung2022DiffusionPS, Chung2022ImprovingDM, Chung2023DecomposedDS, Chung2022Solving3I}. Sequential Monte Carlo diffusion approaches further improve posterior sampling accuracy and demonstrate strong zero-shot generalization capabilities~\cite{Dou2024DiffusionPS}. Despite these advances, most existing diffusion-based CT reconstruction methods are trained under narrowly defined conditions, typically limited to a single dose level or anatomical region. Such specialization restricts robustness to the diverse noise characteristics, anatomical heterogeneity, and protocol variability encountered in real-world clinical practice~\cite{Chen2023ASCONAS, Gao2025NoiseInspiredDM}. Training separate models for different doses, anatomies, or scanners is computationally expensive and does not scale to the diversity of clinical acquisition settings. An overview of the proposed GenDiff framework is illustrated in Figure.~\ref{fig:concept}, highlighting the unified doseand anatomy-aware cold diffusion process with physics-consistent refinement for robust low-dose CT reconstruction.

Low-dose CT reconstruction seeks to recover diagnostically reliable images from measurements acquired under reduced radiation exposure. This problem is ill posed due to severe noise artifacts and information loss. Existing learning-based methods are typically trained for fixed dose levels or specific anatomies. Such specialization limits robustness under realistic clinical variability. Diffusion-based reconstruction has recently shown strong generative capability but most methods remain condition specific and computationally demanding. The central problem addressed in this work is how to design a single diffusion model that generalizes across radiation dose levels and anatomical regions while preserving physical consistency and structural fidelity. To address this challenge we propose a unified framework that integrates dose and anatomy aware conditioning physics-guided refinement and learned structural priors within a generalizable diffusion process. The proposed framework integrates learned generative priors with measurement-domain constraints through iterative refinement and produces reconstructions that are perceptually faithful structurally accurate and physically consistent. Comprehensive experiments on multi-anatomy clinical datasets as well as unseen ultra-low-dose conditions and out-of-distribution phantom and animal data confirm that GenDiff achieves strong empirical generalization across unseen dose levels and two out-of-distribution datasets, without additional retraining or adaptation under the evaluated conditions. The main contributions of this work are summarized as follows:
\begin{itemize}
\item We propose a generalizable diffusion framework that jointly models continuous radiation dose and anatomical region within a single reconstruction network.
\item We introduce a physics-regularized Structural Prior Refinement Module that preserves fine anatomical structures while enforcing consistency with the CT forward model.
\item We demonstrate robust generalization through extensive evaluation on multi-anatomy clinical data unseen ultra-low-dose settings and out-of-distribution phantom and animal datasets without retraining.
\end{itemize}
\section{Method}\label{sec2}
\begin{figure}[!ht]
    \centering
    \includegraphics[width=1\linewidth]{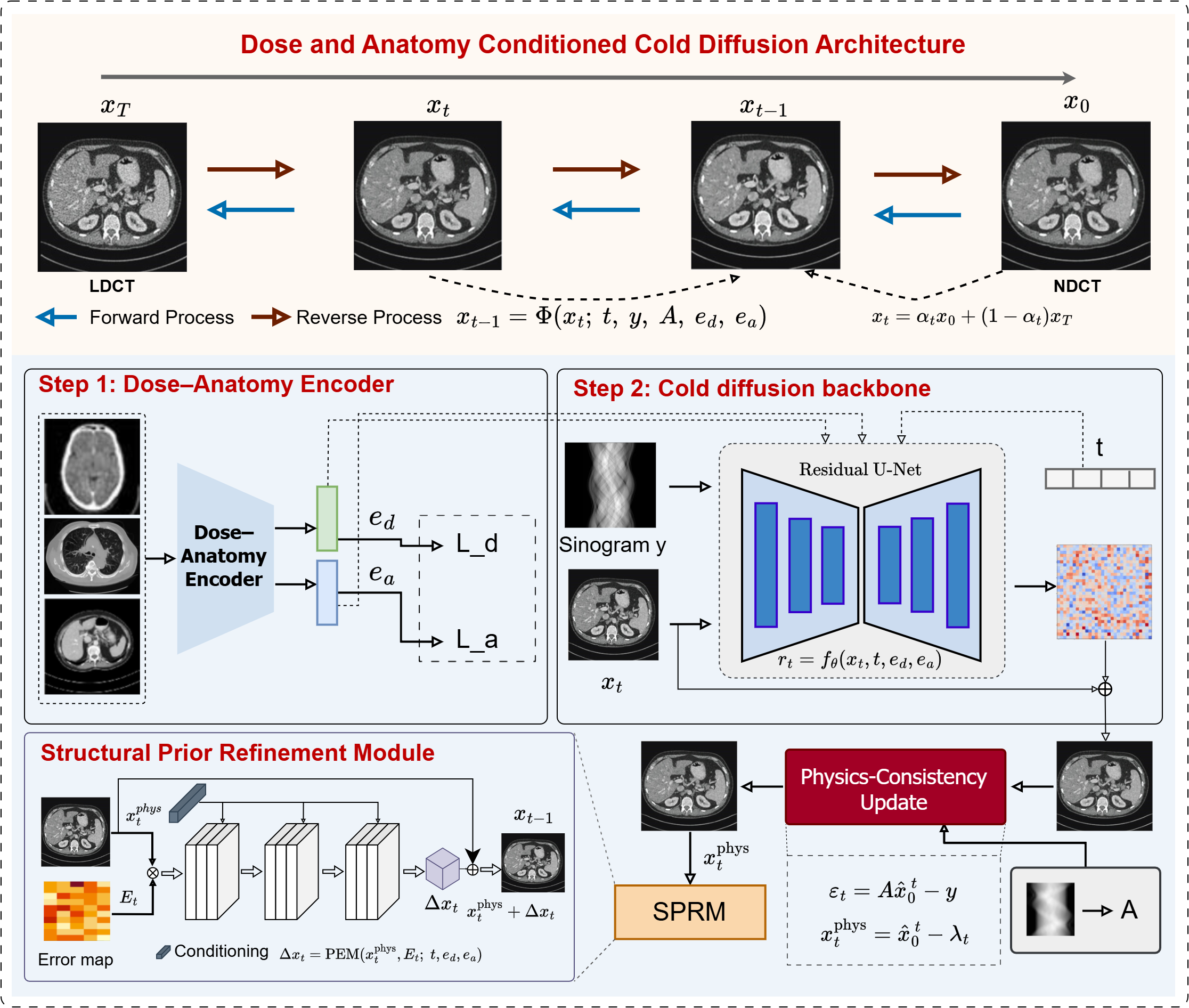}
     \caption{Overview of GenDiff for generalizable low-dose CT reconstruction. GenDiff performs doseand anatomy conditioned cold diffusion to iteratively refine an LDCT input $x_T$ into an NDCT-like reconstruction $x_0$. A pretrained dose-anatomy encoder provides embeddings ($e_d$, $e_a$) to condition a residual U-Net backbone. Each reverse step combines structural prior refinement (SPRM) with a physics-consistent update enforcing data fidelity using the sinogram $y$ and system operator $A$.}
    \label{fig:overall_architecture}
\end{figure}

An overview of the proposed GenDiff framework is shown in Figure.~\ref{fig:overall_architecture}. The method formulates low-dose CT reconstruction as a doseand anatomy-conditioned cold diffusion process, in which a degraded input image is progressively refined through a sequence of deterministic reverse steps. The framework integrates continuous conditioning, structural prior refinement, and physics-consistent updates within each diffusion iteration.
\subsection{Problem Formulation}
The reconstruction of high-quality CT images from noisy, dose-limited projection measurements is fundamentally governed by the physics of X-ray acquisition. In clinical CT, the measured sinogram $y \in \mathbb{R}^{M}$ is related to the underlying attenuation distribution $x \in \mathbb{R}^{H \times W}$ through the discrete Radon transform, expressed as
\vspace{-2mm}
\begin{equation}
    y = A x + n_d,
\end{equation}
where $A$ denotes the system matrix describing the forward projection operator and $n_d$ represents dose-dependent measurement noise arising from reduced photon statistics at radiation dose level $d \in (0,1]$. As $d$ decreases the variance of $n_d$ increases thereby degrading the quality of the reconstructed low-dose CT (LDCT) image $x_T^{d,a}$ while the corresponding high-dose NDCT image $x_0$ serves as an approximation to the clinical reference $x^\ast$. The reconstruction task therefore seeks to recover an estimate $\hat{x}$ that remains faithful to $x_0$ while satisfying the physical constraints imposed by the forward operator.To explicitly incorporate anatomical and dose variability we formulate the reconstruction process as a conditional mapping
\begin{equation}
    \hat{x} = f_\theta(x_T^{d,a}, d, a, y, A),
\end{equation}
where $f_\theta$ denotes a diffusion-based reconstruction operator parameterized by $\theta$ and $(d,a)$ encode continuous radiation dose and anatomical region, respectively. An effective reconstruction model must jointly satisfy three desiderata: (i) \emph{image-level fidelity}, requiring $\hat{x} \approx x_0$; (ii) \emph{physics consistency}, ensuring that $A \hat{x} \approx y$; and (iii) \emph{robust conditioning}, enabling stable performance across a continuous range of doses and across anatomies in $\mathcal{A}=\{\text{abdomen},\text{chest},\text{head}\}$. GenDiff addresses this objective through iterative refinement that integrates dose-anatomy aware diffusion updates with physics-regularized corrections. To promote unified learning across domains, a single parameter set $\theta$ is jointly optimized over LDCT training samples drawn from multiple anatomies and mid-dose levels $d \in \{0.10,0.25\}$, while evaluation includes both seen and unseen dose conditions (5\%, 10\%, 25\%, 50\%). The global optimization objective balances image accuracy, data-fidelity enforcement, and structural preservation through the composite loss
\begin{equation}
    \mathcal{L}(\theta)
    = w_1 \| \hat{x} - x_0 \|_1
    + w_2 \| A \hat{x} - y \|_2^2
    + w_3 \| \nabla \hat{x} - \nabla x_0 \|_1,
\end{equation}
which constrains the reconstruction to be simultaneously consistent with the measurement physics, robust to variations in dose and anatomy, and faithful to the structural characteristics of the underlying NDCT image.
\subsection{GenDiff Overview}
GenDiff is designed as a unified reconstruction framework that integrates dose-anatomy aware conditioning, image-domain cold diffusion, and physics-guided iterative refinement into a single coherent generative process. Given an LDCT reconstruction $x_T^{d,a}$ obtained at dose level $d$ and anatomical region $a$, together with measurement data $y$ and projection operator $A$, the model produces an estimate $\hat{x}$ by iteratively traversing the reverse diffusion trajectory. Each reverse step consists of three components: a denoising update learned by the diffusion backbone, a data-consistency update based on the measured projections, and a proximal-style correction produced by the Structural Prior Refinement Module (SPRM). Together these updates encourage agreement with both the NDCT image prior and the CT forward model. At diffusion timestep \(t\), the network takes the degraded sample \(x_t\) generated by the cold diffusion process and applies a conditional denoising operator \(f_\theta(\cdot)\), with conditioning on dose level and anatomical information. Specifically, the backbone predicts a residual term
\begin{equation}
    r_t = f_\theta(x_t, t, e_d, e_a),
\end{equation}
where $e_d$ and $e_a$ denote the learned embeddings associated with dose level $d$ and anatomical region $a$, respectively. This yields a tentative clean estimate
\begin{equation}
    \hat{x}_0^{\,t} = x_t + r_t,
\end{equation}
which serves as the input to a physics-based refinement step. To enforce compatibility with the measurement model, GenDiff computes the projection residual 
\begin{equation}
    \varepsilon_t = A \hat{x}_0^{\,t} - y,
\end{equation}
and performs a gradient-based data-fidelity update of the form
\begin{equation}
    x^{t,\mathrm{phys}} 
    = \hat{x}_0^{\,t} - \lambda_t A^\top \varepsilon_t,
    \label{eq:phys_corr}
\end{equation}
where $\lambda_t > 0$ is a learnable, timestep-dependent step size. This update corresponds to a single iteration of proximal or gradient descent on the data-consistency term $\| A x - y \|_2^2$ and ensures that the estimate remains aligned with the measurement physics. Following the physics-consistency step, the model applies a learned correction via the Structural Prior Refinement Module (SPRM), which acts as a data-driven proximal operator capturing structural priors and spatially adaptive corrections. The SPRM produces a refinement term
\begin{equation}
    \Delta x_t = \mathrm{SPRM}(x^{t,\mathrm{phys}}, t, e_d, e_a),
\end{equation}
resulting in the updated reverse-diffusion state
\begin{equation}
    x_{t-1} = x^{t,\mathrm{phys}} + \Delta x_t.
\end{equation}
Through the combination of dose-anatomy-aware conditioning, deterministic cold diffusion, and physics-guided optimization, the model iteratively progresses toward an NDCT-quality reconstruction. After $T$ reverse steps, the final reconstruction is obtained as $\hat{x} = x_0$, representing the model’s estimate of the underlying high-dose image. The synergy between learned priors and physically grounded consistency updates enables GenDiff to generalize effectively across dose levels, anatomical regions, and acquisition domains.
\subsection{Model Components}
GenDiff is built from four principal modules: one, a Dose-Anatomy Encoder that embeds acquisition-specific attributes into a compact conditioning space; two, an image-domain cold diffusion backbone that predicts denoising residuals across multiple scales; three, a contextual error modulation mechanism that provides spatially adaptive refinement cues; and four, a physics-regularized Structural Prior Refinement Module (SPRM) that enforces data fidelity while capturing high-level structural priors. These components operate jointly within each reverse diffusion step to produce progressively refined reconstructions. 
\textbf{Dose-Anatomy Encoder:} To enable generalization across dose levels and anatomical regions, GenDiff incorporates a conditioning encoder that maps the continuous dose value $d$ and anatomy identifier $a$ into learned embeddings. Given an input LDCT slice $x_T^{d,a}$, the encoder produces
\begin{equation}
    (e_d, e_a) = \mathrm{Enc}(x_T^{d,a}, d, a),
\end{equation}
where $e_d \in \mathbb{R}^{C}$ is a continuous dose embedding and $e_a \in \mathbb{R}^{C}$ is an anatomy specific feature vector. These embeddings guide the diffusion backbone through adaptive normalization layers, ensuring that the denoising dynamics are conditioned on acquisition attributes. We pretrain the encoder with combination of dose regression loss, dose ranking constraints, and supervised contrastive anatomy classification to impose structured relationships within the embedding space.
\textbf{Cold Diffusion Backbone:} The main denoising module of GenDiff is a multi-scale residual U-Net-like architecture operating under the cold diffusion formulation. At timestep $t$, the backbone receives the degraded image $x_t$ and predicts a residual
\begin{equation}
    r_t = f_\theta(x_t, t, e_d, e_a),
\end{equation}
yielding a clean reconstruction $\hat{x}_0^{\,t} = x_t + r_t$. The backbone leverages adaptive normalization conditioned on concatenated timestep and dose embeddings, enabling the network to modulate its denoising behavior as a function of the forward diffusion level and acquisition dose. Multi-scale attention blocks capture local CT texture patterns and long-range anatomical dependencies, while skip connections keep fine grained spatial detail needed for accurate reconstruction.
\textbf{Contextual Error Modulation:} To enhance spatial adaptiveness and guide the model toward regions requiring stronger regularization, GenDiff computes a contextual difficulty map derived from discrepancies between current estimates and measured projections. Let the projection residual at timestep $t$ be
\begin{equation}
    \varepsilon_t = A \hat{x}_0^{\,t} - y,
\end{equation}
and let $B_t = A^\top \varepsilon_t$ denote its backprojection to image space. Gradient and intensity discrepancies are similarly defined as
\begin{equation}
    G_t = \bigl\lvert \nabla \hat{x}_0^{\,t} - \nabla x_t \bigr\rvert, 
    \qquad 
    I_t = \bigl\lvert \hat{x}_0^{\,t} - x_t \bigr\rvert.
\end{equation}
A lightweight convolutional module aggregates these cues to form
\begin{equation}
    E_t = \sigma\!\left( \mathrm{Conv}\bigl([I_t,\, G_t,\, B_t]\bigr) \right),
\end{equation}
where $\sigma(\cdot)$ denotes a sigmoid activation and $[\cdot]$ represents channel concatenation. The resulting map $E_t \in [0,1]^{H \times W}$ highlights spatial regions where the current estimate exhibits higher uncertainty or inconsistency, and is incorporated into both the diffusion backbone and SPRM to facilitate targeted refinement.
\textbf{Physics-Regularized Structural Prior Refinement Module:} Following the physics update in each reverse diffusion step, GenDiff applies a learned proximal correction to capture structural priors not explicitly modeled by the projection operator. Given the physics-corrected estimate $x^{t,\mathrm{phys}}$, the SPRM computes a refinement term
\begin{equation}
    \Delta x_t = \mathrm{SPRM}(x^{t,\mathrm{phys}}, t, e_d, e_a, E_t),
\end{equation}
where conditioning on $(t,e_d,e_a)$ enables temporal and acquisition-aware modulation, and $E_t$ provides spatial adaptiveness. The update is then applied as
\begin{equation}
    x_{t-1} = x^{t,\mathrm{phys}} + \Delta x_t.
\end{equation}
The SPRM therefore functions as a data-driven regularizer analogous to a learned proximal operator, complementing the physics update while preserving anatomical structure and suppressing dose-dependent noise. Its integration with the diffusion backbone and error modulation ensures that each reverse step achieves a balanced correction guided by both learned priors and measurement fidelity.
\subsection{Training Strategy}
The GenDiff framework is trained in two stages. First, we pretrain the Dose-Anatomy Encoder to obtain robust conditioning embeddings. Second, joint optimization of the diffusion backbone, physics-regularized update, and Structural Prior Refinement Module (SPRM). This two-step process keeps training stable across multi-dose and multi-anatomy settings while preserving the physical consistency necessary for high-quality CT reconstruction.
\textbf{Stage 1: Pretraining of Dose-Anatomy Encoder:} Given an LDCT slice $x_T^{d,a}$ with associated dose level $d$ and anatomy label $a$, the encoder produces embeddings $(e_d, e_a) = \mathrm{Enc}(x_T^{d,a}, d, a)$. The encoder is trained using a combination of regression, ranking, and contrastive losses. Dose regression encourages accurate mapping between physical dose and latent representation:
\begin{equation}
    \mathcal{L}_{\mathrm{dose}} 
    = \bigl\| d_{\mathrm{pred}} - d \bigr\|_2^2,
\end{equation}
where $d_{\mathrm{pred}}$ is the predicted dose value. A monotonicity-preserving ranking loss is imposed for pairs $(i,j)$ with $d_i < d_j$:
\begin{equation}
    \mathcal{L}_{\mathrm{rank}}
    = \sum_{i,j} \max\bigl(0,\, 1 - (d_j - d_i)\langle e_{d,j} - e_{d,i},\, u \rangle \bigr),
\end{equation}
with $u$ a learnable direction vector enforcing ordered dose separation in embedding space. To distinguish anatomies, a supervised contrastive loss is applied:
\begin{equation}
    \mathcal{L}_{\mathrm{anat}}
    = - \sum_{i \in \mathcal{B}} 
      \log \frac{
        \sum_{j \in \mathcal{P}(i)} \exp\left( \langle e_{a,i}, e_{a,j} \rangle / \tau \right)
      }{
        \sum_{k \in \mathcal{B}} \exp\left( \langle e_{a,i}, e_{a,k} \rangle / \tau \right)
      },
\end{equation}
where $\mathcal{B}$ is the batch index set, $\mathcal{P}(i)$ denotes samples sharing anatomy class with $i$, and $\tau$ is a temperature parameter. The encoder is optimized using
\begin{equation}
    \mathcal{L}_{\mathrm{Enc}} 
    = \mathcal{L}_{\mathrm{dose}}
    + \lambda_{\mathrm{rank}} \mathcal{L}_{\mathrm{rank}}
    + \lambda_{\mathrm{anat}} \mathcal{L}_{\mathrm{anat}},
\end{equation}
and its parameters are either frozen or fine-tuned during Stage~2 with reduced learning rate.
\textbf{Stage 2: Joint Optimization of Diffusion and Physics Modules:} For each training pair $(x_0, x_T^{d,a})$, a timestep $t$ is sampled uniformly from $\{1,\dots,T\}$, and the degradation state $x_t$ is computed via the cold diffusion process. The diffusion backbone predicts a residual $r_t$, producing a tentative clean estimate $\hat{x}_0^{\,t} = x_t + r_t$. A physics-based update is then applied,
\begin{equation}
    x^{t,\mathrm{phys}} 
    = \hat{x}_0^{\,t} - \lambda_t A^\top (A \hat{x}_0^{\,t} - y),
\end{equation}
followed by the SPRM refinement
\begin{equation}
    x_{t-1} = x^{t,\mathrm{phys}} + \mathrm{SPRM}(x^{t,\mathrm{phys}}, t, e_d, e_a, E_t).
\end{equation}

The training objective integrates image fidelity, physics consistency, and structural accuracy:
\begin{equation}
    \mathcal{L}_{\mathrm{img}}
    = \bigl\| \hat{x}_0^{\,t} - x_0 \bigr\|_1,
\end{equation}
\vspace{-7mm}
\begin{equation}
    \mathcal{L}_{\mathrm{phys}}
    = \bigl\| A x_{t-1} - y \bigr\|_2^2,
\end{equation}
\vspace{-7mm}
\begin{equation}
    \mathcal{L}_{\mathrm{grad}}
    = \bigl\| \nabla x_{t-1} - \nabla x_0 \bigr\|_1.
\end{equation}
The complete loss for Stage~2 is thus
\begin{equation}
    \mathcal{L}_{\mathrm{Stage2}}
    = w_1 \mathcal{L}_{\mathrm{img}}
    + w_2 \mathcal{L}_{\mathrm{phys}}
    + w_3 \mathcal{L}_{\mathrm{grad}}.
\end{equation}
\textbf{Optimization Details:} Optimization is performed using the AdamW optimizer with cosine learning-rate decay and gradient clipping for stability. Training uses jointly across all anatomies $a \in \{\text{abdomen},\text{chest},\text{head}\}$ at mid-dose levels $d \in \{0.10, 0.25\}$, enabling the model to learn priors that work across different doses and anatomies. At inference, the reverse diffusion process is executed for $T$ steps, produces the final reconstruction $\hat{x} = x_0$. This training setup allows GenDiff to generalize effectively to unseen dose levels and out-of-domain data.
\begin{figure}[!ht]
    \centering
    \includegraphics[width=1\linewidth]{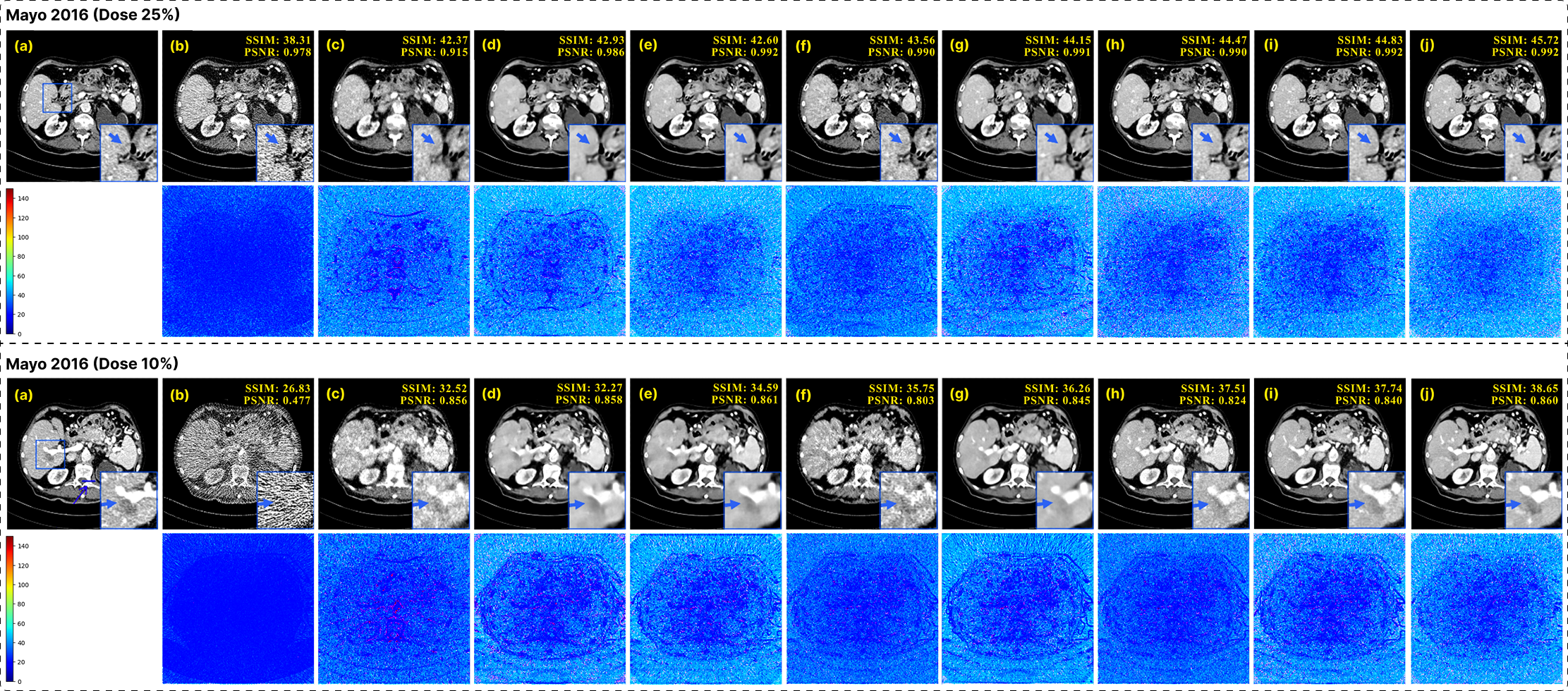}
    \caption{Qualitative comparison on the Mayo-2016 dataset at seen dose levels. Representative abdominal CT slices reconstructed at 25\% and 10\% dose are shown for the NDCT reference, LDCT input, and competing diffusion-based methods, along with the proposed GenDiff. Zoomed-in regions (blue boxes) highlight local structural details and noise characteristics.}
    \label{fig:mayo2016}
\end{figure}
\section{Experiments}\label{sec3}
This section describes the datasets, baseline methods, implementation details, and evaluation protocols used to assess the performance and generalization capability of GenDiff. Experiments were designed to evaluate reconstruction quality across multiple anatomies, dose levels, and acquisition domains, with particular emphasis on cross-dose generalization and robustness to domain shifts. All learning-based baselines were trained or evaluated following their original published protocols. When multi-dose or multi-anatomy training was not supported by the original method, the best-performing reported configuration was used.
\subsection{Datasets}
\textbf{Mayo-2016 Dataset\footnote{\url{https://www.aapm.org/grandchallenge/lowdosect/}, DOI: \url{https://doi.org/10.1002/mp.12345}}:} 
The Mayo Clinic Low-Dose CT (LDCT) Dataset released in 2016 was provided for the NIH-AAPM-Mayo Clinic Low-Dose CT Grand Challenge and has become a standard benchmark for LDCT reconstruction and denoising research \cite{mccollough2017low}. The dataset contains abdominal CT scans from 10 anonymized patients, where normal-dose CT (NDCT) images are available along with simulated low-dose counterparts. The LDCT images are generated through a validated projection-domain noise simulation pipeline, in which NDCT images are forward-projected and corrupted with Poisson and electronic noise to emulate reduced photon counts, followed by filtered back-projection (FBP) reconstruction. In this work, we adopt the commonly used 25\% dose level for evaluation, consistent with prior diffusion-based and deep learning methods. All images are represented with a spatial resolution of $512 \times 512$ pixels. While the Mayo 2016 dataset enables reliable paired quantitative evaluation, it is limited to a single anatomical region and a small cohort, and is therefore mainly used in this study as a controlled benchmark rather than for assessing large-scale generalization.

\textbf{Mayo-2020 Multi-Anatomy Dataset\footnote{\url{https://www.cancerimagingarchive.net/collection/ldct-and-projection-data/}, DOI: \url{https://doi.org/10.1002/mp.14594}}:} 
The Mayo Clinic Low-Dose CT Dataset released in 2020 extends the earlier benchmark by adding multiple anatomical regions and dose levels, making it well suited for evaluating dose-adaptive and generalizable reconstruction methods \cite{moen2021low}. It includes CT scans of the abdomen, chest, and head, with along with normal-dose CT (NDCT) images and multiple simulated low-dose CT (LDCT) versions at different dose levels, typically including 5\%, 10\%, 25\%, and 50\% of the normal dose. Low-dose images are generated using projection-domain simulation with realistic Poisson noise modeling, followed by FBP reconstruction. This produces realistic dose-dependent noise and streak artifacts. In this work, we use the Mayo 2020 dataset to evaluate the dose generalization capability of GenDiff by training the model using only intermediate dose levels (10\% and 25\%) and testing on unseen dose settings, including ultra-low-dose (5\%) cases. This experimental setup follows recent generalized diffusion-based LDCT studies and reflects real clinical scenarios where a single model must work across different radiation dose protocols without retraining.

\textbf{Piglet CT Dataset\footnote{\url{https://doi.org/10.1007/s10278-018-0056-0}}:} 
To evaluate robustness under biological and scanner-induced domain shift, we further employed a Piglet CT dataset~\cite{yi2018sharpness} acquired at a low-dose setting of 10\%. This dataset consists of in vivo animal CT scans exhibiting anatomical structures, tissue compositions, and noise characteristics that differ substantially from the human clinical datasets used during training~\cite{yi2018sharpness}. Differences in body size, organ morphology, and attenuation profiles introduce a challenging out-of-distribution scenario for low-dose CT reconstruction models. GenDiff was applied directly to the Piglet dataset without any retraining or fine-tuning, thereby providing a stringent assessment of cross-domain generalization. Quantitative evaluation was conducted using PSNR, SSIM, and RMSE with respect to available reference reconstructions, while qualitative analysis focused on the preservation of anatomical coherence and suppression of dose-dependent artifacts. This evaluation protocol enables systematic assessment of the model’s robustness to biological variability and acquisition-domain discrepancies beyond human clinical data.

\textbf{Physical Phantom Dataset\footnote{\url{https://www.cancerimagingarchive.net/collection/cc-radiomics-phantom-3/}, DOI: \url{https://doi.org/10.1038/s41598-018-31509-z}}:} 
We evaluated reconstruction performance under acquisition-domain shift using a physical anthropomorphic CT phantom scanned at a radiation dose level of 50\%. The phantom dataset provides a controlled and reproducible acquisition environment with well-defined material inserts and known attenuation properties, enabling reliable assessment of reconstruction fidelity and accuracy~\cite{ger2018comprehensive, zhovannik2019learning}. Unlike simulated or clinical datasets, phantom measurements reflect real scanner physics, including detector response, beam hardening, and system noise, which are not explicitly modeled during training. As with the Piglet dataset, GenDiff was tested on the phantom data without any retraining or fine-tuning. Quantitative evaluation included PSNR, SSIM, and RMSE to assess both image quality and quantitative accuracy. This dataset therefore serves as a complementary domain-shift benchmark for probing the model's ability to generalize across acquisition hardware and physical measurement conditions.
\begin{figure}[!ht]
    \centering
    \includegraphics[width=1\linewidth]{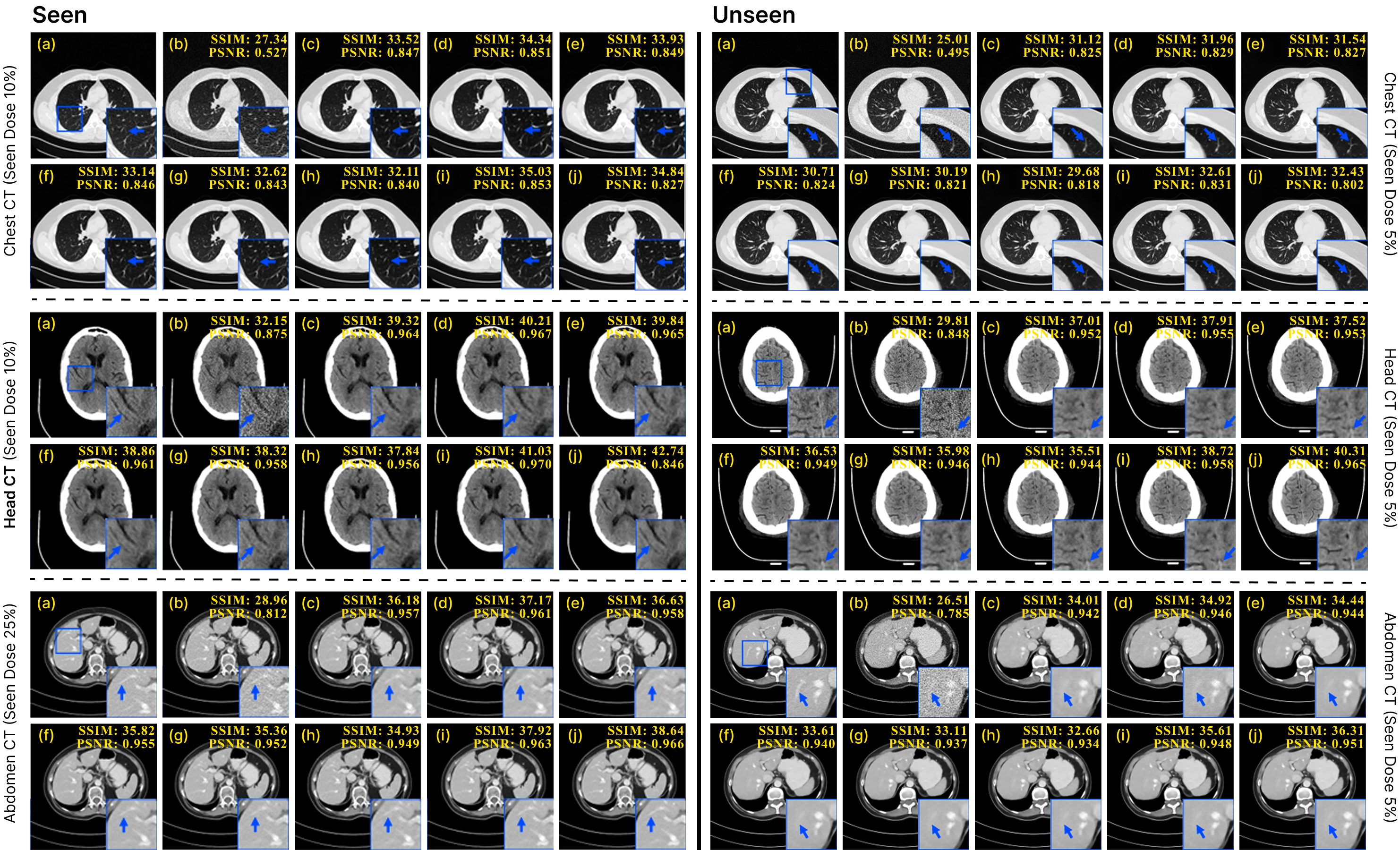}
    \caption{Qualitative comparison on the Mayo-2020 dataset across multiple anatomies and dose levels. Representative axial slices from the chest, head, and abdomen are shown at seen (25\%, 10\%) and unseen ultra-low (5\%) dose settings. From left to right in each group: NDCT reference, LDCT input, representative CNN- and diffusion-based baselines, and the proposed GenDiff. Zoomed-in regions (blue boxes) highlight local anatomical details and artifact patterns.}
    \label{fig:mayo2020}
\end{figure}
\subsection{Data Preprocessing}
To ensure consistency across datasets and reproducible training and evaluation, all CT data were processed using a unified preprocessing pipeline applied across all datasets unless otherwise specified.
\textbf{Slice Extraction:} All volumetric CT scans were converted into 2D axial slices, with each slice treated as an independent sample during training and evaluation.
\textbf{Resizing:} All images were resized to a fixed resolution of $256 \times 256$ pixels to standardize input dimensions and reduce computational cost.
\textbf{Intensity Normalization:} CT intensities were normalized to the range $[0,1]$. For datasets in Hounsfield Units (HU), intensities were linearly scaled after optional clipping to a predefined window to ensure numerical stability and consistency across datasets.
\textbf{Dose Representation:} Each sample was associated with a normalized dose value $d \in (0,1]$. For datasets with discrete dose levels, corresponding values (e.g., $0.25$, $0.10$, $0.05$) were assigned and used as continuous conditioning inputs.
\textbf{Anatomy Labeling:} Each sample was labeled with an anatomical category $a \in \{\text{abdomen}, \text{chest}, \text{head}\}$ to enable multi-anatomy conditioning.
\textbf{Projection Data Handling:} For datasets providing projection (sinogram) data (e.g., Mayo-2020), measurements were incorporated into the physics-consistency module. For datasets without projection data, reconstruction was performed using image-domain inputs only.
\textbf{Data Splits:} Training, validation, and test splits were defined at the patient level to prevent data leakage. For example, the Mayo-2016 dataset was split into 8 training and 2 testing patients. Multi-anatomy datasets were split to ensure coverage across anatomical regions.
\textbf{Consistency Across Datasets:} All preprocessing operations were applied consistently across datasets to ensure that performance differences reflect model behavior rather than preprocessing variations. This unified pipeline is essential for evaluating cross-dose and cross-domain generalization.
\subsection{Baselines}
GenDiff was evaluated against a diverse set of analytical, learning-based, and diffusion-driven reconstruction methods. Filtered back-projection (FBP)~\cite{kak2001principles} served as the analytical baseline. The diffusion-based reconstruction techniques included Diffusion Posterior Sampling (DPS)~\cite{chung2022diffusion}, cold diffusion (ColdDiff)~\cite{bansal2022cold}, contextual error-modulated diffusion (CoreDiff)~\cite{chung2023corediff}, reconstruction-driven diffusion modeling (RDDM)~\cite{wu2023rddm}, a standard Denoising Diffusion Probabilistic Model with 1000 sampling steps (DDPM-1000)~\cite{ho2020ddpm}, and RED-diff~\cite{xia2023reddiff}, a residual encoder-decoder diffusion architecture. Additionally, Noise2Sim~\cite{hendriksen2020noise2sim} was included as a representative self-supervised denoising framework that leverages measurement consistency without requiring paired NDCT targets. Finally, PrideDiff~\cite{prideref} a physics-regularized iterative diffusion framework for LDCT reconstruction was included as the strongest recent diffusion-based baseline, given its state-of-the-art performance on the Mayo benchmarks. Collectively, these baselines encompass state-of-the-art approaches across traditional reconstruction, supervised deep learning, self-supervised denoising, and modern diffusion-based generative modeling, enabling a thorough evaluation of the proposed method. These baselines were selected to cover analytical, supervised, self-supervised, and diffusion-based reconstruction paradigms.
\subsection{Implementation Details} All experiments were implemented in PyTorch and executed on a workstation equipped with four NVIDIA RTX 4090 GPUs. Mixed-precision training was employed to improve memory efficiency. The GenDiff model was trained jointly across multiple anatomies using only mid-dose LDCT data ($d \in \{0.10, 0.25\}$). Unless otherwise stated, the number of reverse diffusion steps was fixed to $T = 20$. 

\textbf{Physics-Consistency Module Operationalization:} The physics-consistency update (Eq.~7) requires access to the sinogram $y$ and the forward projection operator $A$. For datasets that provide projection data-specifically Mayo-2020-we use the discrete Radon transform implemented via the ASTRA Toolbox as the system operator, with $A^\top$ computed as the corresponding filtered backprojection. The timestep-dependent step size $\lambda_t$ is learned end-to-end during Stage~2 training, initialized to $\lambda_t = 0.01$ for all $t$ and optimized jointly with the diffusion backbone and SPRM via the composite loss in Eq.~(26). For datasets where sinogram data are unavailable-namely Mayo-2016, the Piglet dataset, and the Physical Phantom-the physics-consistency update is omitted and reconstruction proceeds using image-domain inputs only. In this image-domain-only mode, the SPRM still receives the physics-corrected estimate $x_t^{\mathrm{phys}}$ replaced by the raw backbone output $\hat{x}_0^t$, preserving the structural refinement pathway. The impact of sinogram availability is evaluated directly, where we report PSNR, SSIM, and RMSE on the Mayo-2020 abdomen subset under both the full physics-guided and image-domain-only configurations. 

\textbf{Training and Inference Cost:} Training times varied between 48-72 hours depending on batch size, and peak memory usage per GPU remained below 20~GB. At inference, reconstruction of a $512 \times 512$ slice required approximately 0.15-0.25 seconds with $T = 20$ reverse steps on a single RTX~4090. GenDiff achieves a favourable trade-off: it is substantially faster than DDPM-1000 (which requires 1000 sampling steps) and comparable to CoreDiff and PrideDiff at equivalent step counts, while delivering superior reconstruction quality. For full 3D volumes, inference time scales linearly with slice count, remaining within clinically acceptable bounds for offline reconstruction workflows.
\subsection{Evaluation Metrics and Protocols}
Quantitative evaluation of reconstruction performance was conducted using widely adopted image-quality and physics-consistency metrics. Let $\hat{x}$ denote a reconstructed image and $x_0$ the corresponding NDCT reference. The peak signal-to-noise ratio (PSNR) was computed as
\begin{equation}
    \mathrm{PSNR} = 10 \log_{10} \left( 
        \frac{L^2}{\| \hat{x} - x_0 \|_2^2 / (HW)}
    \right),
\end{equation}
where $L$ is the dynamic range of CT intensities and $H \times W$ the spatial resolution. Structural similarity (SSIM) was used to assess perceptual fidelity through luminance, contrast, and structural comparisons:
\begin{equation}
    \mathrm{SSIM}(\hat{x}, x_0) 
    = \frac{(2\mu_{\hat{x}}\mu_{x_0} + C_1)(2\sigma_{\hat{x}x_0} + C_2)}
           {(\mu_{\hat{x}}^2 + \mu_{x_0}^2 + C_1)(\sigma_{\hat{x}}^2 + \sigma_{x_0}^2 + C_2)},
\end{equation}
with local means $\mu_{\cdot}$, variances $\sigma_{\cdot}^2$, covariance $\sigma_{\hat{x}x_0}$, and stabilizing constants $C_1, C_2$. Root-mean-square error (RMSE) was also reported:
\begin{equation}
    \mathrm{RMSE}
    = \sqrt{\frac{1}{HW}\,\| \hat{x} - x_0 \|_2^2}.
\end{equation}
These metrics capture both systematic bias and variability in reconstructed attenuation values. Physics consistency was assessed via the projection-domain residual
\begin{equation}
    \mathcal{R}_{\mathrm{phys}} 
    = \| A \hat{x} - y \|_2^2,
\end{equation}
which reflects adherence to the underlying forward model. Lower values of $\mathcal{R}_{\mathrm{phys}}$ indicate improved compatibility with measured projections and reduced deviation from acquisition physics. All metrics were computed slice-wise and then averaged across test subjects. Evaluation was conducted under two primary protocols: (i) \textit{cross-dose reconstruction}, in which the model trained at mid-dose levels (10\%, 25\%) was tested on both seen (10\%, 25\%) and unseen (5\%, 50\%) dose settings, and (ii) \textit{cross-domain generalization}, in which the trained model was applied directly to Piglet and physical phantom datasets without any retraining or domain adaptation. This protocol enables rigorous assessment of robustness to anatomical variability, dose shifts, and acquisition-domain differences.
\begin{table*}[!ht]
\centering
\caption{Quantitative results (mean $\pm$ std) under 25\% and 10\% dose settings of Mayo 2016. Best results are in \textbf{bold}.}
\label{tab:quant_results}
\setlength{\tabcolsep}{2.5pt}
\renewcommand{\arraystretch}{1.15}
\footnotesize

\begin{tabular}{l ccc ccc}
\toprule
\multirow{2}{*}{Method} 
& \multicolumn{3}{c}{25\% Dose} 
& \multicolumn{3}{c}{10\% Dose} \\
\cmidrule(lr){2-4} \cmidrule(lr){5-7}
& PSNR $\uparrow$ & SSIM $\uparrow$ & RMSE $\downarrow$ 
& PSNR $\uparrow$ & SSIM $\uparrow$ & RMSE $\downarrow$ \\
\midrule

FBP 
& 38.31 $\pm$ 1.55 & 0.9783 $\pm$ 0.0080 & 0.0049 $\pm$ 0.0012 
& 26.83 $\pm$ 2.26 & 0.4773 $\pm$ 0.0987 & 0.0530 $\pm$ 0.0148 \\

RDDM* 
& 42.37 $\pm$ 6.87 & 0.9152 $\pm$ 0.0760 & 0.0100 $\pm$ 0.0062 
& 32.52 $\pm$ 1.75 & 0.8560 $\pm$ 0.0545 & 0.0152 $\pm$ 0.0042 \\

ColdDiffusion* 
& 42.93 $\pm$ 1.33 & 0.9869 $\pm$ 0.0075 & 0.0036 $\pm$ 0.0016 
& 32.27 $\pm$ 1.60 & 0.8586 $\pm$ 0.0540 & 0.0156 $\pm$ 0.0036 \\

DDPM-1000* 
& 42.60 $\pm$ 1.09 & 0.9926 $\pm$ 0.0022 & 0.0030 $\pm$ 0.0011 
& 34.59 $\pm$ 1.68 & 0.8615 $\pm$ 0.0522 & 0.0149 $\pm$ 0.0039 \\

RED-diff* 
& 43.56 $\pm$ 0.94 & 0.9905 $\pm$ 0.0026 & 0.0033 $\pm$ 0.0008 
& 35.75 $\pm$ 1.58 & 0.8032 $\pm$ 0.0641 & 0.0186 $\pm$ 0.0035 \\

CoreDiff* 
& 44.15 $\pm$ 1.12 & 0.9910 $\pm$ 0.0030 & 0.0031 $\pm$ 0.0006 
& 36.26 $\pm$ 0.60 & 0.8459 $\pm$ 0.0527 & 0.0163 $\pm$ 0.0042 \\

Noise2Sim* 
& 44.47 $\pm$ 1.38 & 0.9906 $\pm$ 0.0033 & 0.0032 $\pm$ 0.0007 
& 37.51 $\pm$ 1.89 & 0.8246 $\pm$ 0.0637 & 0.0175 $\pm$ 0.0037 \\

PrideDiff* 
& 44.83 $\pm$ 1.08 & 0.9923 $\pm$ 0.0023 & 0.0031 $\pm$ 0.0005 
& 37.74 $\pm$ 1.74 & 0.8408 $\pm$ 0.0597 & 0.0166 $\pm$ 0.0034 \\

\midrule
\textbf{GenDiff (ours)} 
& \textbf{45.72 $\pm$ 1.17} 
& \textbf{0.9928 $\pm$ 0.0023} 
& \textbf{0.0029 $\pm$ 0.0004} 
& \textbf{38.65 $\pm$ 1.78} 
& \textbf{0.8603 $\pm$ 0.0536} 
& \textbf{0.0150 $\pm$ 0.0032} \\
\bottomrule

\end{tabular}
\end{table*}
\section{Results}
\subsection{Performance at Seen Dose Levels}
We next evaluate the reconstruction performance of GenDiff under matched training conditions using the Mayo-2016 dataset at the seen dose levels of 25\% and 10\%. Quantitative results are reported in Table~\ref{tab:quant_results}, where GenDiff is compared against representative diffusion-based reconstruction methods, including ColdDiffusion, DDPM-1000, CoreDiff, RDDM, and PrideDiff. At both dose levels, GenDiff achieves the highest PSNR and SSIM values while yielding the lowest RMSE among all competing methods. In particular, at the 25\% dose level, GenDiff attains a PSNR of $45.72\,\mathrm{dB}$ and an RMSE of $0.0029$, outperforming the strongest diffusion baselines by a clear margin. Similar trends are observed at the more challenging 10\% dose setting, where GenDiff maintains superior reconstruction accuracy with improved PSNR and SSIM compared to prior diffusion-based approaches. Beyond mean performance, GenDiff also exhibits reduced standard deviation across test samples, indicating more stable and reliable reconstructions under matched acquisition conditions. The qualitative comparisons in Figure~\ref{fig:mayo2016} further corroborate the quantitative findings. At both 25\% and 10\% dose levels, GenDiff produces reconstructions with effective noise suppression and substantially reduced streak artifacts, while preserving fine anatomical structures and soft-tissue details. In contrast, competing diffusion-based methods either retain residual noise patterns or exhibit localized over-smoothing, particularly in low-contrast regions. The zoomed-in regions highlight that GenDiff better preserves structural boundaries and texture consistency, closely resembling the NDCT reference.
\begin{table*}[!ht]
\centering
\caption{Quantitative reconstruction performance on the Mayo-2020 dataset across different anatomies and dose levels. Results are reported as mean $\pm$ standard deviation for PSNR, SSIM, and RMSE. Best results are highlighted in \textbf{bold}. * indicates statistically significant difference from the best-performing method ($p < 0.05$, Wilcoxon signed-rank test).}
\label{tab:mayo2020_seen_unseen_multiDose}

\setlength{\tabcolsep}{2pt}
\renewcommand{\arraystretch}{1.08}
\scriptsize

\resizebox{\textwidth}{!}{%
\begin{tabular}{llccccccccc}
\toprule
\textbf{Anatomy} & \textbf{Method}
& \multicolumn{3}{c}{\textbf{25\% (Seen)}}
& \multicolumn{3}{c}{\textbf{10\% (Seen)}}
& \multicolumn{3}{c}{\textbf{5\% (Unseen)}} \\
\cmidrule(lr){3-5} \cmidrule(lr){6-8} \cmidrule(lr){9-11}
& 
& PSNR$\uparrow$ & SSIM$\uparrow$ & RMSE$\downarrow$
& PSNR$\uparrow$ & SSIM$\uparrow$ & RMSE$\downarrow$
& PSNR$\uparrow$ & SSIM$\uparrow$ & RMSE$\downarrow$ \\
\midrule

\multirow{9}{*}{\textbf{Abdomen}}
& FBP
& 31.42$\pm$1.45 & 0.845$\pm$0.010 & 0.032$\pm$0.006
& 28.96$\pm$2.50 & 0.812$\pm$0.012 & 0.039$\pm$0.005
& 26.51$\pm$2.55 & 0.785$\pm$0.014 & 0.045$\pm$0.004 \\

& ColdDiffusion*
& 38.07$\pm$1.30 & 0.964$\pm$0.004 & 0.018$\pm$0.003
& 36.18$\pm$1.33 & 0.957$\pm$0.005 & 0.022$\pm$0.003
& 34.01$\pm$1.35 & 0.942$\pm$0.006 & 0.027$\pm$0.009 \\

& CoreDiff*
& 38.94$\pm$1.29 & 0.967$\pm$0.004 & 0.017$\pm$0.004
& 37.17$\pm$1.32 & 0.961$\pm$0.005 & 0.021$\pm$0.006
& 34.92$\pm$1.34 & 0.946$\pm$0.006 & 0.026$\pm$0.005 \\

& RDDM*
& 38.32$\pm$2.30 & 0.965$\pm$0.004 & 0.018$\pm$0.005
& 36.63$\pm$1.33 & 0.958$\pm$0.005 & 0.022$\pm$0.005
& 34.44$\pm$1.36 & 0.944$\pm$0.006 & 0.027$\pm$0.003 \\

& DDPM-1000*
& 37.64$\pm$1.31 & 0.962$\pm$0.005 & 0.019$\pm$0.003
& 35.82$\pm$2.34 & 0.955$\pm$0.006 & 0.023$\pm$0.002
& 33.61$\pm$2.36 & 0.940$\pm$0.007 & 0.028$\pm$0.007 \\

& RED-diff*
& 37.14$\pm$1.32 & 0.959$\pm$0.005 & 0.020$\pm$0.002
& 35.36$\pm$1.35 & 0.952$\pm$0.006 & 0.024$\pm$0.004
& 33.11$\pm$1.38 & 0.937$\pm$0.007 & 0.029$\pm$0.003 \\

& Noise2Sim*
& 36.62$\pm$2.33 & 0.956$\pm$0.005 & 0.021$\pm$0.004
& 34.93$\pm$2.36 & 0.949$\pm$0.006 & 0.025$\pm$0.002
& 32.66$\pm$2.38 & 0.934$\pm$0.007 & 0.030$\pm$0.006 \\

& \textit{PrideDiff}*
& 39.83$\pm$2.28 & 0.970$\pm$0.004 & 0.016$\pm$0.002
& 37.92$\pm$1.31 & 0.963$\pm$0.005 & 0.020$\pm$0.003
& 35.61$\pm$1.34 & 0.948$\pm$0.006 & 0.025$\pm$0.004 \\

& \textbf{GenDiff (Ours)}
& \textbf{40.57$\pm$1.21} & \textbf{0.982$\pm$0.002} & \textbf{0.015$\pm$0.001}
& \textbf{38.64$\pm$1.24} & \textbf{0.966$\pm$0.003} & \textbf{0.019$\pm$0.002}
& \textbf{36.31$\pm$1.27} & \textbf{0.951$\pm$0.004} & \textbf{0.024$\pm$0.003} \\

\midrule

\multirow{9}{*}{\textbf{Chest}}
& FBP
& 29.87$\pm$2.48 & 0.681$\pm$0.012 & 0.036$\pm$0.004
& 27.34$\pm$2.52 & 0.527$\pm$0.014 & 0.042$\pm$0.006
& 25.01$\pm$2.55 & 0.495$\pm$0.016 & 0.048$\pm$0.010 \\

& ColdDiffusion*
& 35.41$\pm$1.30 & 0.857$\pm$0.005 & 0.023$\pm$0.006
& 33.52$\pm$2.33 & 0.847$\pm$0.006 & 0.027$\pm$0.003
& 31.12$\pm$2.36 & 0.825$\pm$0.007 & 0.031$\pm$0.004 \\

& CoreDiff*
& 36.20$\pm$1.29 & 0.860$\pm$0.005 & 0.022$\pm$0.003
& 34.34$\pm$1.32 & 0.851$\pm$0.003 & 0.026$\pm$0.004
& 31.96$\pm$1.35 & 0.829$\pm$0.007 & 0.030$\pm$0.007 \\

& RDDM*
& 35.74$\pm$1.30 & 0.858$\pm$0.005 & 0.022$\pm$0.005
& 33.93$\pm$2.33 & 0.849$\pm$0.004 & 0.026$\pm$0.006
& 31.54$\pm$2.36 & 0.827$\pm$0.007 & 0.030$\pm$0.008 \\

& DDPM-1000*
& 35.03$\pm$2.31 & 0.855$\pm$0.006 & 0.023$\pm$0.002
& 33.14$\pm$1.34 & 0.846$\pm$0.006 & 0.027$\pm$0.002
& 30.71$\pm$1.36 & 0.824$\pm$0.007 & 0.031$\pm$0.005 \\

& RED-diff*
& 34.45$\pm$2.32 & 0.852$\pm$0.006 & 0.024$\pm$0.004
& 32.62$\pm$2.35 & 0.843$\pm$0.005 & 0.028$\pm$0.005
& 30.19$\pm$2.38 & 0.821$\pm$0.007 & 0.032$\pm$0.006 \\

& Noise2Sim*
& 33.91$\pm$1.33 & 0.849$\pm$0.006 & 0.025$\pm$0.002
& 32.11$\pm$2.36 & 0.840$\pm$0.004 & 0.029$\pm$0.003
& 29.68$\pm$2.38 & 0.818$\pm$0.007 & 0.033$\pm$0.009 \\

& \textit{PrideDiff}*
& 36.95$\pm$2.29 & 0.862$\pm$0.005 & 0.021$\pm$0.003
& 35.03$\pm$1.32 & 0.853$\pm$0.004 & 0.025$\pm$0.002
& 32.61$\pm$1.35 & 0.831$\pm$0.007 & 0.029$\pm$0.004 \\

& \textbf{GenDiff (Ours)}
& \textbf{38.13$\pm$2.22} & \textbf{0.892$\pm$0.003} & \textbf{0.017$\pm$0.001}
& \textbf{34.84$\pm$1.25} & \textbf{0.827$\pm$0.002} & \textbf{0.024$\pm$0.002}
& \textbf{32.43$\pm$1.28} & \textbf{0.802$\pm$0.006} & \textbf{0.029$\pm$0.003} \\

\midrule

\multirow{9}{*}{\textbf{Head}}
& FBP
& 34.62$\pm$1.42 & 0.902$\pm$0.009 & 0.028$\pm$0.004
& 32.15$\pm$2.47 & 0.875$\pm$0.011 & 0.034$\pm$0.007
& 29.81$\pm$2.50 & 0.848$\pm$0.013 & 0.039$\pm$0.006 \\

& ColdDiffusion*
& 41.44$\pm$2.28 & 0.969$\pm$0.004 & 0.015$\pm$0.002
& 39.32$\pm$2.31 & 0.964$\pm$0.005 & 0.018$\pm$0.002
& 37.01$\pm$2.34 & 0.952$\pm$0.006 & 0.022$\pm$0.009 \\

& CoreDiff*
& 42.35$\pm$1.27 & 0.972$\pm$0.003 & 0.014$\pm$0.005
& 40.21$\pm$1.30 & 0.967$\pm$0.004 & 0.017$\pm$0.005
& 37.91$\pm$1.33 & 0.955$\pm$0.006 & 0.021$\pm$0.003 \\

& RDDM*
& 41.83$\pm$1.28 & 0.970$\pm$0.004 & 0.015$\pm$0.002
& 39.84$\pm$2.31 & 0.965$\pm$0.005 & 0.018$\pm$0.004
& 37.52$\pm$2.34 & 0.953$\pm$0.006 & 0.022$\pm$0.007 \\

& DDPM-1000*
& 40.94$\pm$1.29 & 0.967$\pm$0.004 & 0.016$\pm$0.006
& 38.86$\pm$2.32 & 0.961$\pm$0.005 & 0.019$\pm$0.006
& 36.53$\pm$2.35 & 0.949$\pm$0.006 & 0.023$\pm$0.005 \\

& RED-diff*
& 40.34$\pm$2.30 & 0.964$\pm$0.005 & 0.017$\pm$0.004
& 38.32$\pm$1.33 & 0.958$\pm$0.005 & 0.020$\pm$0.003
& 35.98$\pm$1.36 & 0.946$\pm$0.006 & 0.024$\pm$0.008 \\

& Noise2Sim*
& 39.83$\pm$2.31 & 0.962$\pm$0.005 & 0.018$\pm$0.005
& 37.84$\pm$1.34 & 0.956$\pm$0.006 & 0.021$\pm$0.002
& 35.51$\pm$1.37 & 0.944$\pm$0.007 & 0.025$\pm$0.006 \\

& \textit{PrideDiff}*
& 43.11$\pm$1.26 & 0.975$\pm$0.003 & 0.013$\pm$0.003
& 41.03$\pm$1.29 & 0.970$\pm$0.004 & 0.016$\pm$0.004
& 38.72$\pm$1.32 & 0.958$\pm$0.005 & 0.020$\pm$0.007 \\

& \textbf{GenDiff (Ours)}
& \textbf{44.51$\pm$1.20} & \textbf{0.993$\pm$0.002} & \textbf{0.012$\pm$0.001}
& \textbf{42.74$\pm$2.23} & \textbf{0.977$\pm$0.003} & \textbf{0.015$\pm$0.002}
& \textbf{40.31$\pm$2.26} & \textbf{0.965$\pm$0.004} & \textbf{0.019$\pm$0.003} \\
\bottomrule
\end{tabular}%
}
\end{table*}
\subsection{Overall Performance Across Anatomies and Doses}
\begin{figure*}[!ht]
    \centering
    \includegraphics[width=\textwidth]{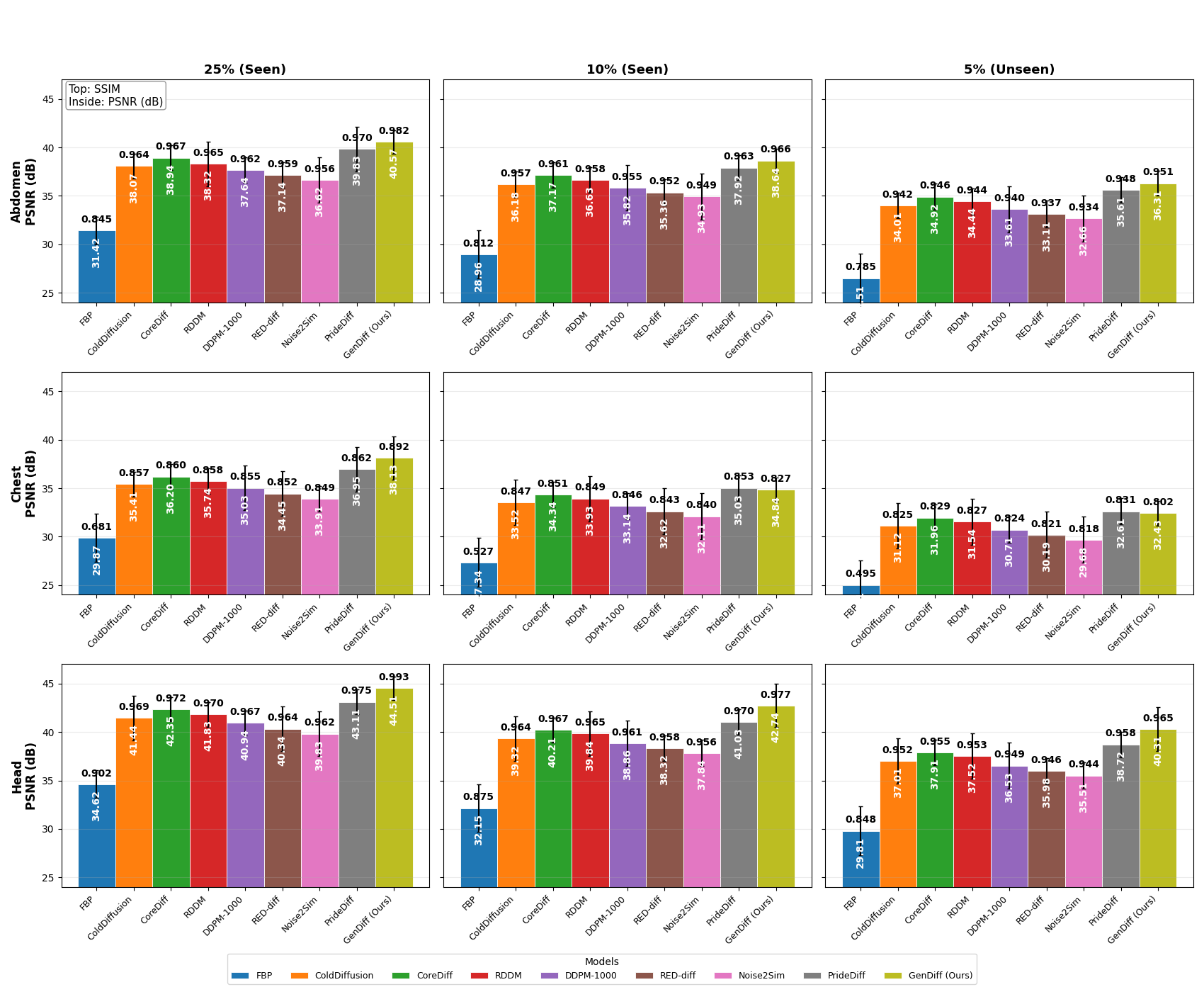}
    \caption{Quantitative comparison on the Mayo-2020 dataset across anatomies and dose levels. Results are shown for abdomen (top), chest (middle), and head (bottom) at seen doses (25\%, 10\%) and an unseen ultra-low dose (5\%). Bars report PSNR (dB) and annotations report SSIM. GenDiff achieves the best performance across all anatomies, with particularly strong gains at 5\%, demonstrating robust cross-dose generalization.}
    \label{fig:quantitative_psnr_ssim}
\end{figure*}

We present an overall quantitative and qualitative evaluation of GenDiff on the Mayo-2020 dataset, which comprises abdomen, chest, and head CT images reconstructed under multiple radiation dose levels. Table~\ref{tab:mayo2020_seen_unseen_multiDose} summarizes the quantitative reconstruction performance at the seen dose levels (25\% and 10\%) and the unseen ultra-low dose level (5\%), while representative qualitative results are shown in Figure~\ref{fig:mayo2020}. GenDiff consistently achieves the best performance across all three anatomies and dose settings in terms of PSNR and SSIM, while achieving the lowest RMSE compared to baselines. At the seen dose levels of 25\% and 10\%, GenDiff shows clear and stable improvements over strong diffusion competitors such as CoreDiff, RDDM, and PrideDiff, indicating that the proposed unified framework does not compromise reconstruction accuracy under matched training conditions. Importantly, the performance gains are observed uniformly across abdomen, chest, and head anatomies, highlighting the effectiveness of the dose-anatomy-aware conditioning in handling heterogeneous anatomical structures. As shown in figure.~\ref{fig:quantitative_psnr_ssim}, GenDiff achieves the highest PSNR and SSIM across all anatomies and dose levels. At the unseen 5\% dose, GenDiff improves PSNR by up to 1-2 dB over prior diffusion-based methods and maintains higher SSIM values (0.951 vs. 0.948 for abdomen and 0.965 vs. 0.958 for head), showing stronger performance under very low-dose conditions.
More notably, GenDiff maintains a strong performance advantage at the unseen 5\% dose level, where reconstruction becomes substantially more challenging due to severe quantum noise and streak artifacts. While all methods experience performance degradation as the dose decreases, GenDiff exhibits a markedly smaller drop in PSNR and SSIM across all anatomies, demonstrating robust generalization beyond the training dose distribution. This behavior suggests that GenDiff learns a continuous, dose-aware reconstruction process rather than overfitting to discrete noise characteristics associated with specific dose levels. The qualitative comparisons in Figure~\ref{fig:mayo2020} further corroborate the quantitative findings. GenDiff consistently produces cleaner reconstructions with reduced noise and artifacts while preserving fine anatomical details across both seen and unseen dose conditions. The visual results reveal smooth and coherent structural transitions across dose levels, without the over-smoothing or residual artifacts observed in competing methods. Overall, the results in Table~\ref{tab:mayo2020_seen_unseen_multiDose} and Figure~\ref{fig:mayo2020} demonstrate that GenDiff achieves stable dose-response behavior and strong generalization capability across anatomies and radiation dose levels using a single unified model.
\begin{table*}[!ht]
\centering
\caption{Quantitative comparison results on Phantom (50\% dose) and Piglet (10\% dose) CT reconstruction (mean $\pm$ std).}
\label{tab:combined_results}
\resizebox{\textwidth}{!}{
\begin{tabular}{c | ccc | ccc}
\hline
\multirow{2}{*}{Method} 
& \multicolumn{3}{c|}{\textbf{Phantom (50\% dose)}} 
& \multicolumn{3}{c}{\textbf{Piglet (10\% dose)}} \\
\cline{2-7}
& PSNR (dB) & SSIM & RMSE 
& PSNR (dB) & SSIM & RMSE \\
\hline
FBP 
& $37.47 \pm 0.25$ & $0.9449 \pm 0.0024$ & $0.027 \pm 0.005$
& $32.02 \pm 0.33$ & $0.9011 \pm 0.0056$ & $0.037 \pm 0.006$ \\

ColdDiffusion* 
& $40.47 \pm 0.22$ & $0.9558 \pm 0.0020$ & $0.019 \pm 0.002$
& $34.52 \pm 0.19$ & $0.9923 \pm 0.0020$ & $0.034 \pm 0.003$ \\

RED-diff*
& $40.51 \pm 0.23$ & $0.9428 \pm 0.0025$ & $0.018 \pm 0.004$
& $34.85 \pm 0.22$ & $0.9836 \pm 0.0031$ & $0.035 \pm 0.005$ \\

RDDM* 
& $40.63 \pm 0.23$ & $0.9428 \pm 0.0025$ & $0.019 \pm 0.002$
& $34.35 \pm 0.25$ & $0.9237 \pm 0.0047$ & $0.033 \pm 0.004$ \\

DDPM-1000* 
& $41.72 \pm 0.20$ & $0.9547 \pm 0.0021$ & $0.016 \pm 0.005$
& $35.02 \pm 0.24$ & $0.9439 \pm 0.0048$ & $0.031 \pm 0.003$ \\

CoreDiff*  
& $41.36 \pm 0.21$ & $0.9446 \pm 0.0022$ & $0.017 \pm 0.003$
& $35.23 \pm 0.27$ & $0.9321 \pm 0.0050$ & $0.034 \pm 0.005$ \\

Noise2Sim* 
& $43.62 \pm 0.17$ & $0.9723 \pm 0.0015$ & $0.013 \pm 0.001$
& $35.72 \pm 0.18$ & $0.9520 \pm 0.0038$ & $0.029 \pm 0.002$ \\

\textit{PrideDiff}* 
& $44.84 \pm 0.15$ & $0.9728 \pm 0.0013$ & $0.012 \pm 0.002$
& $36.33 \pm 0.16$ & $0.9533 \pm 0.0035$ & $0.028 \pm 0.003$ \\

\hline
\textbf{GenDiff (Ours)} 
& $\mathbf{45.09 \pm 0.14}$ & $\mathbf{0.9770 \pm 0.0012}$ & $\mathbf{0.011 \pm 0.001}$
& $\mathbf{36.85 \pm 0.17}$ & $\mathbf{0.9621 \pm 0.0025}$ & $\mathbf{0.027 \pm 0.001}$ \\

\hline
\end{tabular}
}
\end{table*}
\begin{figure}[!ht]
    \centering
    \includegraphics[width=1\linewidth]{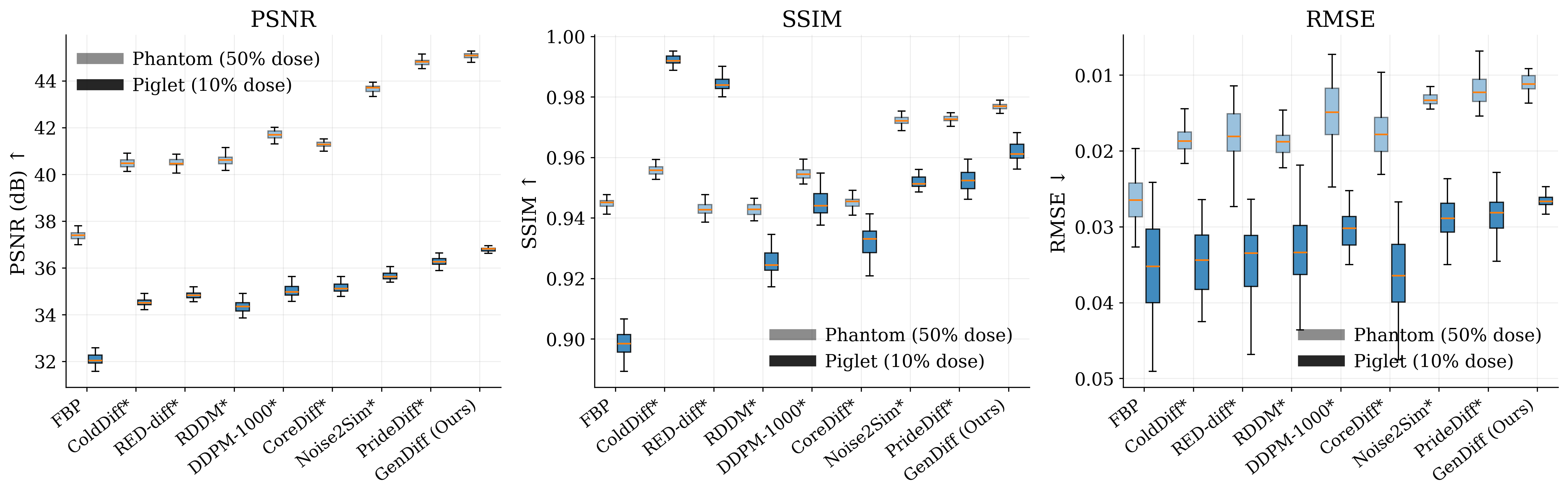}
    \caption{Cross-domain performance distribution on Phantom and Piglet CT datasets. Boxplots of PSNR, SSIM, and RMSE for Phantom (50\% dose) and Piglet (10\% dose) reconstructions across different methods.}
    \label{fig:cross_domain_boxplots}
\end{figure}
\subsection{Robustness to Domain Shift}
\begin{figure}[!ht]
    \centering
    \includegraphics[width=1\linewidth]{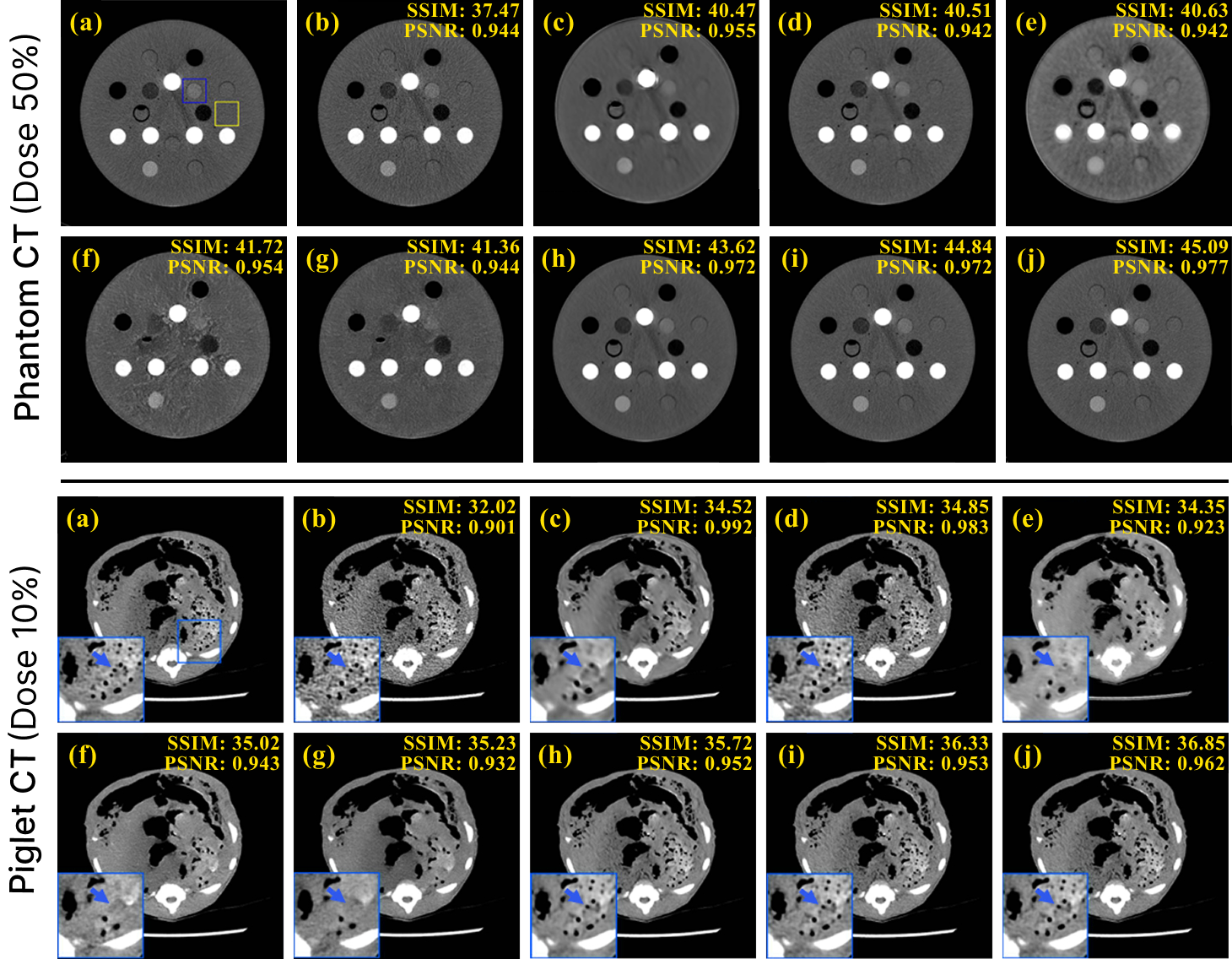}
    \caption{Qualitative results on out-of-distribution datasets. Representative reconstructions from the physical phantom (50\% dose) and Piglet CT (10\% dose) are shown for competing methods and the proposed GenDiff. Zoomed-in regions highlight noise suppression, artifact reduction, and preservation of structural boundaries under domain shift.}
    \label{fig:generalization}
\end{figure}

We further evaluate the robustness of GenDiff under domain shift using two out-of-distribution test sets a physical anthropomorphic phantom scanned at 50\% dose and a Piglet CT dataset acquired at 10\% dose. Importantly GenDiff is applied to both datasets without any retraining or fine-tuning or domain adaptation which provides a stringent assessment of generalization across scanners protocols and anatomical characteristics. Quantitative results are summarized in Table~\ref{tab:combined_results}. Figure~\ref{fig:cross_domain_boxplots} further visualizes the performance distributions across methods for both datasets. As reported in Table~\ref{tab:combined_results} GenDiff consistently achieves the best quantitative performance on both datasets in terms of PSNR and SSIM while yielding the lowest or comparable RMSE relative to competing methods. On the physical phantom dataset GenDiff attains the highest PSNR and SSIM with reduced variance indicating improved reconstruction fidelity and stable performance across repeated measurements. The distributional trends shown in Figure~\ref{fig:cross_domain_boxplots} further highlight the robustness of GenDiff under domain shift. In particular GenDiff exhibits consistently tighter performance distributions across Phantom and Piglet datasets which reflects reduced sensitivity to acquisition variability. The phantom experiments also enable reliable assessment of accuracy where GenDiff demonstrates lower reconstruction error and variance compared to diffusion based baselines reflecting stronger adherence to physical attenuation properties. On the Piglet dataset which exhibits anatomical structures and noise characteristics distinct from the Mayo data GenDiff similarly outperforms existing methods confirming robustness to biological and scanner induced variations. Representative qualitative comparisons are shown in Figure~\ref{fig:generalization}. The qualitative results further support these quantitative findings. For phantom data GenDiff effectively suppresses noise and streak artifacts while preserving sharp boundaries of high contrast inserts and fine structural details. For Piglet CT images GenDiff maintains coherent anatomical structures and reduces artifact contamination more effectively than competing diffusion based methods. The zoomed in regions highlight GenDiff ability to preserve texture consistency and edge definition under heterogeneous acquisition conditions.

\begin{table*}[!ht]
\centering
\caption{Ablation study on the Mayo-2020 dataset for abdomen and head CT images at 5\% dose and chest CT images at 10\% dose. The effects of dose-anatomy conditioning, physics-consistency updates, the structural prior refinement module (SPRM), and contextual error modulation are evaluated. Best results are highlighted in \textbf{bold}.}
\label{tab:ablation_combined}

\setlength{\tabcolsep}{2pt}
\renewcommand{\arraystretch}{1.1}
\footnotesize

\resizebox{\textwidth}{!}{%
\begin{tabular}{ccccc ccc ccc ccc}
\toprule
\multicolumn{5}{c}{\textbf{Components}} 
& \multicolumn{3}{c}{\textbf{Abdomen (5\% Dose)}} 
& \multicolumn{3}{c}{\textbf{Chest (10\% Dose)}} 
& \multicolumn{3}{c}{\textbf{Head (5\% Dose)}} \\
\cmidrule(lr){1-5} \cmidrule(lr){6-8} \cmidrule(lr){9-11} \cmidrule(lr){12-14}
\textbf{Dose} & \textbf{Anat} & \textbf{Phys} & \textbf{SPRM} & \textbf{Err}
& \textbf{PSNR}$\uparrow$ & \textbf{SSIM}$\uparrow$ & \textbf{RMSE}$\downarrow$
& \textbf{PSNR}$\uparrow$ & \textbf{SSIM}$\uparrow$ & \textbf{RMSE}$\downarrow$
& \textbf{PSNR}$\uparrow$ & \textbf{SSIM}$\uparrow$ & \textbf{RMSE}$\downarrow$ \\
\midrule

\checkmark & \checkmark & \checkmark & \checkmark & \checkmark
& \textbf{36.29} & \textbf{0.954} & \textbf{0.025}
& \textbf{34.86} & \textbf{0.825} & \textbf{0.022}
& \textbf{40.30} & \textbf{0.967} & \textbf{0.021} \\

\checkmark & \checkmark & \checkmark & \checkmark & $\times$
& 35.82 & 0.946 & 0.026
& 34.31 & 0.816 & 0.025
& 39.68 & 0.938 & 0.023 \\

\checkmark & \checkmark & \checkmark & $\times$ & $\times$
& 34.14 & 0.939 & 0.028
& 33.72 & 0.814 & 0.028
& 38.96 & 0.948 & 0.023 \\

\checkmark & \checkmark & $\times$ & $\times$ & $\times$
& 33.02 & 0.928 & 0.031
& 32.65 & 0.801 & 0.032
& 37.85 & 0.944 & 0.026 \\

$\times$ & $\times$ & \checkmark & \checkmark & \checkmark
& 35.71 & 0.946 & 0.027
& 34.18 & 0.819 & 0.025
& 39.75 & 0.953 & 0.022 \\

$\times$ & $\times$ & $\times$ & $\times$ & $\times$
& 33.61 & 0.921 & 0.036
& 32.21 & 0.795 & 0.033
& 37.12 & 0.936 & 0.028 \\

\bottomrule
\end{tabular}%
}
\end{table*}

\subsection{Ablation Study}
\label{sec:ablation}
We conduct a comprehensive ablation study to quantify the contribution of individual components in the proposed framework, with quantitative results summarized in Table~\ref{tab:ablation_combined}. Experiments are performed on the Mayo-2020 dataset across three anatomies and dose settings: abdomen and head CT images at an unseen ultra-low dose level (5\%) and chest CT images at a moderate dose level (10\%). The full model, which jointly incorporates dose-anatomy conditioning, physics-consistency updates, the Structural Prior Refinement Module (SPRM), and contextual error modulation, consistently achieves the best performance across all anatomies and metrics. In contrast, removing dose-anatomy conditioning or physics-consistency enforcement results in substantial degradation in PSNR and SSIM and increased RMSE, with the most pronounced performance drop observed at the unseen 5\% dose level, underscoring the importance of acquisition-aware conditioning and explicit physics guidance for robust generalization under severe noise.

\begin{figure}[!ht]
    \centering
    \includegraphics[width=1\linewidth]{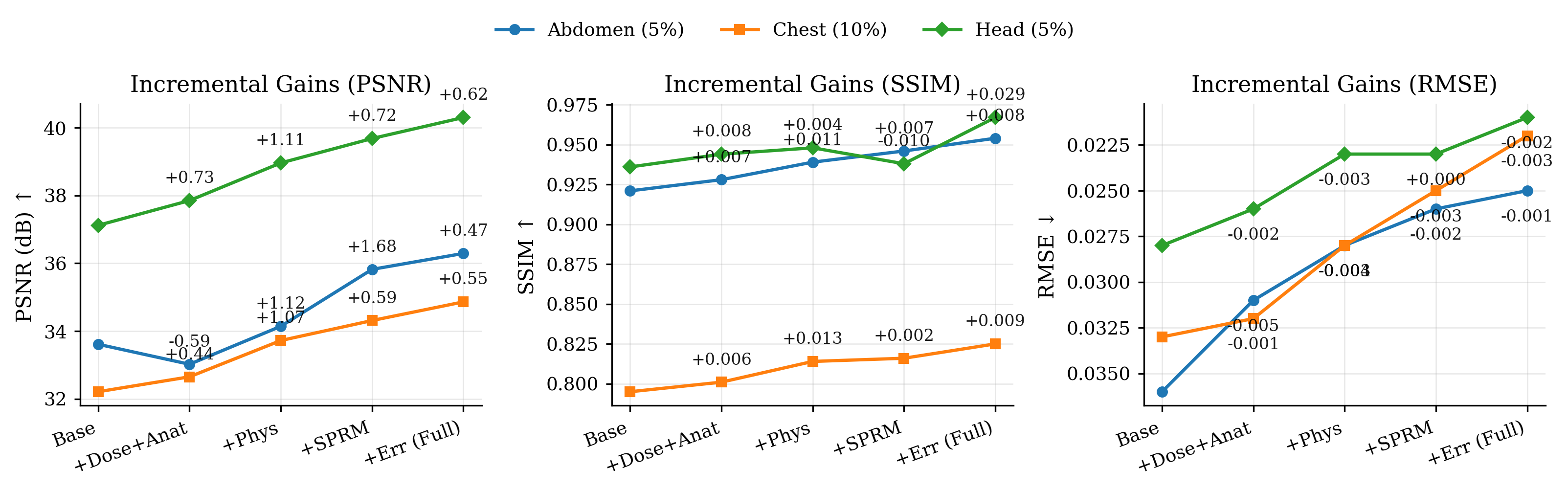}
    \caption{Incremental gains in PSNR, SSIM, and RMSE obtained by progressively adding model components in the ablation study for abdomen (5\%), chest (10\%), and head (5\%) CT images.}
    \label{fig:ablation_incremental_gain}
\end{figure}

To further elucidate the role of each component, Figure~\ref{fig:ablation_incremental_gain} visualizes the incremental performance gains obtained by progressively adding model components, reported in terms of PSNR, SSIM, and RMSE. Across all anatomies, incorporating physics-consistency updates yields the largest single-step improvement, indicating that enforcing measurement fidelity is the dominant factor in improving reconstruction accuracy. The inclusion of SPRM provides additional gains by enhancing structural preservation, while contextual error modulation delivers consistent but smaller improvements, particularly in reducing RMSE through spatially adaptive refinement. Notably, the incremental trends are consistent across abdomen, chest, and head anatomies, demonstrating that each component contributes additively and synergistically to reconstruction quality. Together, the table- and figure-based analyses confirm that the unified integration of conditioning, physics-based updates, and learned structural priors is essential for achieving stable and accurate CT reconstruction across anatomies and varying dose levels.
\section{Discussion and Future Work}
\label{sec:discussion}
\subsection{Discussion of Results}
This work demonstrates that a unified physics- and dose-anatomy-aware diffusion framework can achieve strong generalization across radiation dose levels, anatomical regions, and acquisition domains using a single model trained on limited mid-dose data. The proposed GenDiff framework consistently preserves reconstruction fidelity at unseen ultra-low dose levels, including 5\%, while maintaining stable performance across abdomen, chest, and head anatomies. Moreover, the integration of physics-consistency updates improves adherence to the CT forward model, resulting in reconstructions that are not only perceptually accurate but also physically plausible. The robustness observed under domain shift, including evaluation on physical phantom and animal datasets, further indicates that the proposed method captures transferable priors beyond dataset-specific characteristics. Compared to prior deep learning approaches that rely on dose-specific or anatomy-specific training, GenDiff avoids the need for maintaining multiple specialized models and mitigates overfitting to discrete acquisition settings. In contrast to physics-free diffusion or denoising approaches, which may generate visually appealing but measurement-inconsistent results, the proposed cold diffusion formulation explicitly enforces projection-domain consistency at each reverse step. The use of contextual error modulation further enhances spatial adaptiveness, enabling targeted refinement in regions affected by severe noise or artifacts. Collectively, these design choices highlight the advantage of jointly integrating dose-anatomy conditioning, physics-based updates, and learned structural priors within a unified reconstruction framework.
\subsection{Limitations and Future Directions}
Despite its strengths, this study has several limitations. First, some low-dose CT data used in training and evaluation are generated through dose-dependent simulation rather than direct acquisition, which may not fully capture real-world scanner noise characteristics. Second, although robustness is demonstrated across multiple datasets, the evaluation remains limited to a small number of scanner platforms and acquisition protocols. Third, the current implementation operates on 2D slices, which does not fully exploit volumetric correlations present in three-dimensional CT data. In addition, while inference is substantially faster than classical diffusion models, the iterative sampling process still incurs higher computational cost compared to single-pass reconstruction networks.

Future work will focus on extending the proposed framework to fully 3D and 4D CT reconstruction to better leverage spatial and temporal consistency. Large-scale multi-center and multi-vendor clinical studies will be pursued to further validate robustness and generalizability in real-world settings. Another promising direction is the integration of reconstruction with downstream clinical tasks, such as detection or segmentation, within a joint learning framework. Finally, accelerating inference through reduced sampling steps, progressive distillation, or alternative deterministic solvers remains an important avenue for improving clinical practicality.
\section{Conclusion}\label{sec4}
In this work, we propose GenDiff, a unified physics- and dose-anatomy-aware diffusion framework for generalizable low-dose CT reconstruction. By integrating deterministic cold diffusion with explicit physics-consistency updates and structured conditioning on radiation dose and anatomical region, GenDiff enables a single model to robustly reconstruct high-quality CT images across multiple anatomies, unseen ultra-low dose levels, and acquisition domains. Experiments on multi-anatomy clinical datasets, as well as out-of-distribution phantom and animal data, demonstrate that GenDiff outperforms recent CNN and diffusion-based methods in terms of reconstruction quality, preserves anatomical structures better, and maintains physics consistency. The model also generalizes well to new dose levels and domains without retraining. These results suggest that jointly embedding acquisition-aware conditioning and physics-guided iterative refinement within a diffusion framework can help address common challenges of dose variability, anatomical heterogeneity, and domain shift in low-dose CT reconstruction.
\section*{Acknowledgements}
The authors used ChatGPT~(\cite{openai2025chatgpt}) for general purposes such as grammar refinement, structural organization, and documentation formatting. All scientific content, data processing, and results were independently verified and approved by the authors.
\section*{Code Availability}
The implementation of GenDiff used in this study is publicly available. The source code can be accessed through GitHub at: 
\href{https://github.com/imamahasane/GenDiff}{https://github.com/imamahasane/GenDiff}. An archived and citable version of the code has been deposited in Zenodo and is available at: \href{https://doi.org/10.5281/zenodo.19393226}{https://doi.org/10.5281/zenodo.19393226}.

\bibliography{sample}

\begin{thebibliography}{}

\bibitem[Abdun~Noor et~al., 2025]{abdun2025geglunet}
Abdun~Noor, A., Ahasan, M.~I., Khan, M.~A., and Yang, G. (2025).
\newblock Geglunet: Structural retinal vessel segmentation via attention-gated
  geglu and contrastive supervision.
\newblock In {\em Chinese Conference on Pattern Recognition and Computer Vision
  (PRCV)}, pages 494--507. Springer.

\bibitem[Attivissimo et~al., 2010]{Attivissimo2010ATT}
Attivissimo, F., Cavone, G., Lanzolla, A. M.~L., and Spadavecchia, M. (2010).
\newblock A technique to improve the image quality in computer tomography.
\newblock {\em IEEE Transactions on Instrumentation and Measurement},
  59:1251--1257.

\bibitem[Bansal et~al., 2022a]{Bansal2022ColdDI}
Bansal, A., Borgnia, E., Chu, H.-M., Li, J., Kazemi, H., Huang, F., Goldblum,
  M., Geiping, J., and Goldstein, T. (2022a).
\newblock Cold diffusion: Inverting arbitrary image transforms without noise.
\newblock {\em ArXiv}, abs/2208.09392.

\bibitem[Bansal et~al., 2022b]{bansal2022cold}
Bansal, A., Borgnia, E., and Grover, A. (2022b).
\newblock Cold diffusion: Inverting arbitrary image transforms without noise.
\newblock In {\em Advances in Neural Information Processing Systems (NeurIPS)}.

\bibitem[Camacho-Mondragon et~al., 2025]{camacho2025clinical}
Camacho-Mondragon, C.~G., Ibarrola-Pe{\~n}a, J.~C., Lira-Lozano, D.,
  Jerjes-Sanchez, C., De~la Pena-Almaguer, E., and Paredes-Vazquez, J.~G.
  (2025).
\newblock Clinical applications of cardiac computed tomography: A focused
  review for the clinical cardiologists.
\newblock {\em Journal of Cardiovascular Development and Disease}, 12(10):375.

\bibitem[Chambolle and Lions, 1997]{Chambolle1997ImageRV}
Chambolle, A. and Lions, P.-L. (1997).
\newblock Image recovery via total variation minimization and related problems.
\newblock {\em Numerische Mathematik}, 76:167--188.

\bibitem[Chen et~al., 2017a]{Chen2017LEARNLE}
Chen, H., Zhang, Y., Chen, Y., Zhang, J., hua Zhang, W., Sun, H., Lv, Y., Liao,
  P., Zhou, J., and Wang, G. (2017a).
\newblock Learn: Learned experts’ assessment-based reconstruction network for
  sparse-data ct.
\newblock {\em IEEE Transactions on Medical Imaging}, 37:1333--1347.

\bibitem[Chen et~al., 2017b]{Chen2017LowDoseCW}
Chen, H., Zhang, Y., Kalra, M.~K., Lin, F., Chen, Y., Liao, P., Zhou, J., and
  Wang, G. (2017b).
\newblock Low-dose ct with a residual encoder-decoder convolutional neural
  network.
\newblock {\em IEEE Transactions on Medical Imaging}, 36:2524--2535.

\bibitem[Chen et~al., 2023]{Chen2023ASCONAS}
Chen, Z., Gao, Q., Zhang, Y., and Shan, H. (2023).
\newblock Ascon: Anatomy-aware supervised contrastive learning framework for
  low-dose ct denoising.
\newblock {\em ArXiv}, abs/2307.12225.

\bibitem[Chung et~al., 2022a]{Chung2022DiffusionPS}
Chung, H., Kim, J., McCann, M.~T., Klasky, M.~L., and Ye, J.~C. (2022a).
\newblock Diffusion posterior sampling for general noisy inverse problems.
\newblock {\em ArXiv}, abs/2209.14687.

\bibitem[Chung et~al., 2023a]{chung2022diffusion}
Chung, H., Kim, J., Monga, V., and Ye, J.~C. (2023a).
\newblock Diffusion posterior sampling for general noisy inverse problems.
\newblock In {\em International Conference on Learning Representations (ICLR)}.

\bibitem[Chung et~al., 2023b]{Chung2023DecomposedDS}
Chung, H., Lee, S., and Ye, J.-C. (2023b).
\newblock Decomposed diffusion sampler for accelerating large-scale inverse
  problems.
\newblock In {\em International Conference on Learning Representations}.

\bibitem[Chung et~al., 2022b]{Chung2022Solving3I}
Chung, H., Ryu, D., McCann, M.~T., Klasky, M.~L., and Ye, J.~C. (2022b).
\newblock Solving 3d inverse problems using pre-trained 2d diffusion models.
\newblock {\em 2023 IEEE/CVF Conference on Computer Vision and Pattern
  Recognition (CVPR)}, pages 22542--22551.

\bibitem[Chung et~al., 2022c]{Chung2022ImprovingDM}
Chung, H., Sim, B., Ryu, D., and Ye, J.~C. (2022c).
\newblock Improving diffusion models for inverse problems using manifold
  constraints.
\newblock {\em ArXiv}, abs/2206.00941.

\bibitem[Chung et~al., 2023c]{chung2023corediff}
Chung, H., Sim, B., and Ye, J.~C. (2023c).
\newblock Improving diffusion models for inverse problems using manifold
  constraints.
\newblock In {\em International Conference on Machine Learning (ICML)}.

\bibitem[Dhariwal and Nichol, 2021]{Dhariwal2021DiffusionMB}
Dhariwal, P. and Nichol, A. (2021).
\newblock Diffusion models beat gans on image synthesis.
\newblock {\em ArXiv}, abs/2105.05233.

\bibitem[Dou and Song, 2024]{Dou2024DiffusionPS}
Dou, Z. and Song, Y. (2024).
\newblock Diffusion posterior sampling for linear inverse problem solving: A
  filtering perspective.
\newblock In {\em International Conference on Learning Representations}.

\bibitem[Gao et~al., 2025]{Gao2025NoiseInspiredDM}
Gao, Q., Chen, Z., Zeng, D., Zhang, J., Ma, J., and Shan, H. (2025).
\newblock Noise-inspired diffusion model for generalizable low-dose ct
  reconstruction.
\newblock {\em Medical image analysis}, 105.

\bibitem[Gao et~al., 2023]{Gao2023CoreDiffCE}
Gao, Q., Li, Z., Zhang, J., Zhang, Y., and Shan, H. (2023).
\newblock Corediff: Contextual error-modulated generalized diffusion model for
  low-dose ct denoising and generalization.
\newblock {\em IEEE Transactions on Medical Imaging}, 43:745--759.

\bibitem[Gao and Shan, 2022]{Gao2022CoCoDiffAC}
Gao, Q. and Shan, H. (2022).
\newblock Cocodiff: a contextual conditional diffusion model for low-dose ct
  image denoising.
\newblock {\em Developments in X-Ray Tomography XIV}.

\bibitem[Ger et~al., 2018]{ger2018comprehensive}
Ger, R.~B., Zhou, S., Chi, P.-C.~M., Lee, H.~J., Layman, R.~R., Jones, A.~K.,
  Goff, D.~L., Fuller, C.~D., Howell, R.~M., Li, H., et~al. (2018).
\newblock Comprehensive investigation on controlling for ct imaging
  variabilities in radiomics studies.
\newblock {\em Scientific reports}, 8(1):13047.

\bibitem[He et~al., 2019]{He2019OptimizingAP}
He, J., Yang, Y., Wang, Y., Zeng, D., Bian, Z., Zhang, H., Sun, J., Xu, Z., and
  Ma, J. (2019).
\newblock Optimizing a parameterized plug-and-play admm for iterative low-dose
  ct reconstruction.
\newblock {\em IEEE Transactions on Medical Imaging}, 38:371--382.

\bibitem[Hendriksen et~al., 2020]{hendriksen2020noise2sim}
Hendriksen, A.~A., Pelt, D.~M., and Batenburg, K.~J. (2020).
\newblock Noise2sim: Self-supervised denoising from single noisy images.
\newblock {\em IEEE Transactions on Computational Imaging}.

\bibitem[Ho et~al., 2020a]{Ho2020DenoisingDP}
Ho, J., Jain, A., and Abbeel, P. (2020a).
\newblock Denoising diffusion probabilistic models.
\newblock {\em ArXiv}, abs/2006.11239.

\bibitem[Ho et~al., 2020b]{ho2020ddpm}
Ho, J., Jain, A., and Abbeel, P. (2020b).
\newblock Denoising diffusion probabilistic models.
\newblock {\em Advances in Neural Information Processing Systems}.

\bibitem[Huang et~al., 2021]{Huang2021DUGANGA}
Huang, Z., Zhang, J., Zhang, Y., and Shan, H. (2021).
\newblock Du-gan: Generative adversarial networks with dual-domain u-net-based
  discriminators for low-dose ct denoising.
\newblock {\em IEEE Transactions on Instrumentation and Measurement}, 71:1--12.

\bibitem[Humphries et~al., 2019]{Humphries2019ComparisonOD}
Humphries, T., Si, D., Coulter, S., Simms, M., and Xing, R. (2019).
\newblock Comparison of deep learning approaches to low dose ct using low
  intensity and sparse view data.
\newblock In {\em Medical Imaging}.

\bibitem[Kak and Slaney, 2001]{kak2001principles}
Kak, A.~C. and Slaney, M. (2001).
\newblock {\em Principles of Computerized Tomographic Imaging}.
\newblock SIAM.

\bibitem[Kang et~al., 2018]{Kang2018CycleconsistentAD}
Kang, E., Koo, H.~J., Yang, D.~H., Seo, J.~B., and Ye, J.~C. (2018).
\newblock Cycle‐consistent adversarial denoising network for multiphase
  coronary ct angiography.
\newblock {\em Medical Physics}, 46:550–562.

\bibitem[Kim et~al., 2019]{Kim2019APC}
Kim, B., Han, M., Shim, H., and Baek, J. (2019).
\newblock A performance comparison of cnn-based image denoising methods: The
  effect of loss functions on low-dose ct images.
\newblock {\em Medical physics}.

\bibitem[Lu et~al., 2025]{prideref}
Lu, Z., Gao, Q., Wang, T., Yang, Z., Wang, Z., Yu, H., Chen, H., Zhou, J.,
  Shan, H., and Zhang, Y. (2025).
\newblock Pridediff: Physics-regularized generalized diffusion model for ct
  reconstruction.
\newblock {\em IEEE Transactions on Radiation and Plasma Medical Sciences},
  9(2):157--168.

\bibitem[McCollough et~al., 2017]{mccollough2017low}
McCollough, C.~H., Bartley, A.~C., Carter, R.~E., Chen, B., Drees, T.~A.,
  Edwards, P., Holmes~III, D.~R., Huang, A.~E., Khan, F., Leng, S., et~al.
  (2017).
\newblock Low-dose ct for the detection and classification of metastatic liver
  lesions: results of the 2016 low dose ct grand challenge.
\newblock {\em Medical physics}, 44(10):e339--e352.

\bibitem[Moen et~al., 2021]{moen2021low}
Moen, T.~R., Chen, B., Holmes~III, D.~R., Duan, X., Yu, Z., Yu, L., Leng, S.,
  Fletcher, J.~G., and McCollough, C.~H. (2021).
\newblock Low-dose ct image and projection dataset.
\newblock {\em Medical physics}, 48(2):902--911.

\bibitem[Nagare et~al., 2021]{Nagare2021ABL}
Nagare, M., Melnyk, R., Rahman, O., Sauer, K.~D., and Bouman, C.~A. (2021).
\newblock A bias-reducing loss function for ct image denoising.
\newblock {\em ICASSP 2021 - 2021 IEEE International Conference on Acoustics,
  Speech and Signal Processing (ICASSP)}, pages 1175--1179.

\bibitem[Nichol and Dhariwal, 2021]{Nichol2021ImprovedDD}
Nichol, A. and Dhariwal, P. (2021).
\newblock Improved denoising diffusion probabilistic models.
\newblock {\em ArXiv}, abs/2102.09672.

\bibitem[{OpenAI}, 2025]{openai2025chatgpt}
{OpenAI} (2025).
\newblock Chatgpt (gpt-5).
\newblock \url{https://chat.openai.com/}.
\newblock Large language model developed by OpenAI. Accessed: November 11,
  2025.

\bibitem[Shah and Platt, 2008]{Shah2008ALARAIT}
Shah, N. and Platt, S.~L. (2008).
\newblock Alara: is there a cause for alarm? reducing radiation risks from
  computed tomography scanning in children.
\newblock {\em Current Opinion in Pediatrics}, 20:243–247.

\bibitem[Shan et~al., 2018]{Shan2018CompetitivePO}
Shan, H., Padole, A., Homayounieh, F., Kruger, U., Khera, R.~D., Nitiwarangkul,
  C., Kalra, M.~K., and Wang, G. (2018).
\newblock Competitive performance of a modularized deep neural network compared
  to commercial algorithms for low-dose ct image reconstruction.
\newblock {\em Nature Machine Intelligence}, 1:269 -- 276.

\bibitem[Smith-Bindman et~al., 2009]{smith2009radiation}
Smith-Bindman, R., Lipson, J., Marcus, R., Kim, K.~E., Mahesh, M., Gould, R.,
  Berrington~de Gonzalez, A., and Miglioretti, D.~L. (2009).
\newblock Radiation dose associated with common computed tomography
  examinations and the associated lifetime attributable risk of cancer.
\newblock {\em Archives of Internal Medicine}, 169(22):2078--2086.

\bibitem[Sodickson et~al., 2009]{Sodickson2009RecurrentCC}
Sodickson, A.~D., Baeyens, P.~F., Andriole, K.~P., Prevedello, L.~M., Nawfel,
  R., Hanson, R., and Khorasani, R. (2009).
\newblock Recurrent ct, cumulative radiation exposure, and associated
  radiation-induced cancer risks from ct of adults.
\newblock {\em Radiology}, 251 1.

\bibitem[Sohl-Dickstein et~al., 2015]{SohlDickstein2015DeepUL}
Sohl-Dickstein, J.~N., Weiss, E.~A., Maheswaranathan, N., and Ganguli, S.
  (2015).
\newblock Deep unsupervised learning using nonequilibrium thermodynamics.
\newblock {\em ArXiv}, abs/1503.03585.

\bibitem[Song et~al., 2020]{Song2020ScoreBasedGM}
Song, Y., Sohl-Dickstein, J.~N., Kingma, D.~P., Kumar, A., Ermon, S., and
  Poole, B. (2020).
\newblock Score-based generative modeling through stochastic differential
  equations.
\newblock {\em ArXiv}, abs/2011.13456.

\bibitem[Wang et~al., 2020]{Wang2020DeepLF}
Wang, G., Ye, J.~C., and Man, B.~D. (2020).
\newblock Deep learning for tomographic image reconstruction.
\newblock {\em Nature Machine Intelligence}, 2:737 -- 748.

\bibitem[Wu and et~al., 2023]{wu2023rddm}
Wu, J. and et~al. (2023).
\newblock Reconstruction-driven diffusion models for low-dose ct
  reconstruction.
\newblock {\em IEEE Transactions on Medical Imaging}.

\bibitem[Wu et~al., 2018]{Wu2018NonLocalLC}
Wu, W., Liu, F., Zhang, Y., Wang, Q., and Yu, H. (2018).
\newblock Non-local low-rank cube-based tensor factorization for spectral ct
  reconstruction.
\newblock {\em IEEE Transactions on Medical Imaging}, 38:1079--1093.

\bibitem[Xia and et~al., 2023]{xia2023reddiff}
Xia, B. and et~al. (2023).
\newblock Red-diff: Residual encoder--decoder diffusion models for image
  reconstruction.
\newblock {\em Medical Image Analysis}.

\bibitem[Xia et~al., 2021]{Xia2021CTRW}
Xia, W., Lu, Z., Huang, Y., Liu, Y., Chen, H., Zhou, J., and Zhang, Y. (2021).
\newblock Ct reconstruction with pdf: Parameter-dependent framework for data
  from multiple geometries and dose levels.
\newblock {\em IEEE Transactions on Medical Imaging}, 40:3065--3076.

\bibitem[Xia et~al., 2020]{Xia2020MAGICMA}
Xia, W., Lu, Z., Huang, Y., Shi, Z., Liu, Y., Chen, H., Chen, Y., Zhou, J., and
  Zhang, Y. (2020).
\newblock Magic: Manifold and graph integrative convolutional network for
  low-dose ct reconstruction.
\newblock {\em IEEE Transactions on Medical Imaging}, 40:3459--3472.

\bibitem[Xia et~al., 2022]{Xia2022LowDoseCU}
Xia, W., Lyu, Q., and Wang, G. (2022).
\newblock Low-dose ct using denoising diffusion probabilistic model for 20
  times speedup.
\newblock {\em ArXiv}, abs/2209.15136.

\bibitem[Xu et~al., 2012]{Xu2012LowDoseXC}
Xu, Q., Yu, H., Mou, X., Zhang, L., Hsieh, J., and Wang, G. (2012).
\newblock Low-dose x-ray ct reconstruction via dictionary learning.
\newblock {\em IEEE Transactions on Medical Imaging}, 31:1682--1697.

\bibitem[Yan et~al., 2011]{Yan2011ACS}
Yan, H., Cervi{\~n}o, L.~I., Jia, X., and Jiang, S.~B. (2011).
\newblock A comprehensive study on the relationship between the image quality
  and imaging dose in low-dose cone beam ct.
\newblock {\em Physics in Medicine and Biology}, 57:2063 -- 2080.

\bibitem[Yang et~al., 2022]{Yang2022DiffusionMA}
Yang, L., Zhang, Z., Hong, S., Xu, R., Zhao, Y., Shao, Y., Zhang, W., Yang,
  M.-H., and Cui, B. (2022).
\newblock Diffusion models: A comprehensive survey of methods and applications.
\newblock {\em ACM Computing Surveys}, 56:1 -- 39.

\bibitem[Yang et~al., 2017]{Yang2017LowDoseCI}
Yang, Q., Yan, P., Zhang, Y., Yu, H., Shi, Y., Mou, X., Kalra, M.~K., Zhang,
  Y., Sun, L., and Wang, G. (2017).
\newblock Low-dose ct image denoising using a generative adversarial network
  with wasserstein distance and perceptual loss.
\newblock {\em IEEE Transactions on Medical Imaging}, 37:1348--1357.

\bibitem[Yen et~al., 2022]{Yen2022ColdDF}
Yen, H., Germain, F.~G., Wichern, G., and Roux, J.~L. (2022).
\newblock Cold diffusion for speech enhancement.
\newblock {\em ICASSP 2023 - 2023 IEEE International Conference on Acoustics,
  Speech and Signal Processing (ICASSP)}, pages 1--5.

\bibitem[Yi and Babyn, 2018]{yi2018sharpness}
Yi, X. and Babyn, P. (2018).
\newblock Sharpness-aware low-dose ct denoising using conditional generative
  adversarial network.
\newblock {\em Journal of digital imaging}, 31(5):655--669.

\bibitem[Zhao et~al., 2018]{Zhao2018BiasAG}
Zhao, S., Ren, H., Yuan, A., Song, J., Goodman, N.~D., and Ermon, S. (2018).
\newblock Bias and generalization in deep generative models: An empirical
  study.
\newblock In {\em Neural Information Processing Systems}.

\bibitem[Zhovannik et~al., 2019]{zhovannik2019learning}
Zhovannik, I., Bussink, J., Traverso, A., Shi, Z., Kalendralis, P., Wee, L.,
  Dekker, A., Fijten, R., and Monshouwer, R. (2019).
\newblock Learning from scanners: Bias reduction and feature correction in
  radiomics.
\newblock {\em Clinical and translational radiation oncology}, 19:33--38.

\end{thebibliography}

\end{document}